\documentclass[preprint,5p]{elsarticle}
\usepackage[utf8]{inputenc}
\usepackage{tikz}
\usetikzlibrary{positioning,calc}
\usepackage{natbib}
\usepackage{amsmath}
\usepackage[svgpath=./images/introfig/]{svg}
\usepackage{tummath}
\usepackage{tumcolors}
\usepackage{epflcolors}
\colorlet{tumblue}{EPFLred}
\colorlet{tumorange}{EPFLdark}
\usepackage{csquotes}

\usepackage[backref=page]{hyperref}
\usepackage[capitalize]{cleveref}
\renewcommand{\argmin}{\mathop{\mathrm{argmin}}}
\usepackage{subcaption}
\usepackage{booktabs}
\usepackage{pgfplots}
\usepackage{pgfplotstable}
\usepgfplotslibrary{groupplots, fillbetween, statistics}
\usetikzlibrary{positioning, calc, arrows, arrows.meta, fit, backgrounds, shapes, snakes}
\usepackage{algorithm}
\usepackage{algpseudocode}
\usepackage{siunitx}
\usepackage{lscape}
\usepackage{rotating}
\usepackage{soul}
\usepackage{cancel}

\usepackage{lineno}
\usepackage{lipsum}
\usepackage[most]{tcolorbox}

\newtcolorbox[auto counter]{answer}[3][]{%
	enhanced,frame empty,interior empty,
	colframe=blue,
	borderline west={1pt}{0pt}{green!25!blue},
	coltitle=black,
	boxed title style={boxrule=.4pt,sharp corners},
	colback = white,
	colbacktitle = acier!10,
	title={#2},
	label={#3},
	#1
}

\crefname{appendix}{}{}

\usepackage{graphicx}
\graphicspath{{images/introfig/}}

\makeatletter
\newcommand{\showfontsize}{\f@size{} pt}
\makeatother

\usetikzlibrary{external}
\tikzexternaldisable

\renewcommand{\class}[1]{\textsc{#1}}

\newcommand{\new}[1]{{#1}}
\newcommand{\remove}[1]{}
\newcommand{\rearrange}[1]{#1}

\definecolor{rouge}{RGB}{255,0,0}      % 1
\definecolor{leman}{RGB}{0,167,159}      % 1
\definecolor{grosseille}{RGB}{181,31,31}      % 1
\definecolor{canard}{RGB}{0,116,128}      % 1
\definecolor{montrose}{RGB}{243,152,105}      % 1
\definecolor{perle}{RGB}{202,199,199}      % 1
\definecolor{vertedeau}{RGB}{194,221,176}      % 1
\definecolor{rose}{RGB}{237,110,156}      % rose
\definecolor{acier}{RGB}{79,143,204}      % blue
\definecolor{soufre}{RGB}{251,238,102}      % yellow
\definecolor{carotte}{RGB}{236,102,8}      % yellow
\definecolor{zinzolin}{RGB}{92,36,131}      % purple
\definecolor{chartreuse}{RGB}{200,211,0}      % green
\definecolor{marron}{RGB}{91,52,40}      % brown
\definecolor{ardoise}{RGB}{69,58,76}      % darkblue
\definecolor{taupe}{RGB}{65,61,58}      % darkgray

\definecolor{colororange}{HTML}{E65100} % orange
\definecolor{colordgray}{HTML}{795548} % dark gray for note
\definecolor{colorhgray}{HTML}{212121} % heavy dark gray for normal text
\definecolor{colorgreen}{HTML}{009688} % green
\definecolor{colorlgray}{HTML}{FFFFFF} % FAFAFA background light gray

\colorlet{colorclassone}{tumblue}
\colorlet{colorclasstwo}{tumblack}
\colorlet{colorclassthree}{tumgreen}
\colorlet{colorclassfour}{tumorange}
\colorlet{colorblue}{tumblue}

\colorlet{worsecolor}{tumdiagramdarkred}
\colorlet{bettercolor}{tumblue}

\colorlet{earlinesssixcolor}{tumblack}
\colorlet{earlinesssevencolor}{tumbluedark}
\colorlet{earlinesseightcolor}{tumorange}
\colorlet{earlinessninecolor}{tumbluelight}

\colorlet{indicatorlosscolor}{colorblue}
\colorlet{nlllosscolor}{colorgreen}
\colorlet{linearlosscolor}{colororange}
\colorlet{colortrain}{colorblue}
\colorlet{colorinfer}{colororange}

\definecolor{evalcolor}{HTML}{3F3F3F}
\definecolor{traincolor}{HTML}{B98951}
\definecolor{validcolor}{HTML}{3F4BBE}

\colorlet{colortrain}{tumblue}
\colorlet{colorinfer}{tumblack}

\colorlet{earlinesscolor}{tumblue}
\colorlet{accuracycolor}{tumorange}

\colorlet{stdcolor}{tumbluelight}
\colorlet{mediancolor}{tumorange}
\colorlet{meancolor}{tumblue}

\colorlet{b1color}{tumblack}%tumdiagramaubergine
\colorlet{b9color}{tumblack}%tumblack
\colorlet{b10color}{tumblack}%tumblue

\colorlet{b2color}{tumblue}%tumdiagramnavyblue
\colorlet{b3color}{tumblue}%tumdiagramturquoise
\colorlet{b4color}{tumblue}%tumdiagramgreen

\colorlet{b5color}{tumdiagramred}%tumdiagramlimegreen
\colorlet{b6color}{tumdiagramred}%tumdiagramyellow
\colorlet{b7color}{tumdiagramred}%tumdiagramsand
\colorlet{b8color}{tumdiagramred}%tumdiagramredorange
\colorlet{b8Acolor}{tumdiagramred}%tumdiagramred

\colorlet{b11color}{tumorange}%tumdiagramdarkred
\colorlet{b12color}{tumorange}%tumorange

\colorlet{epsilon0color}{tumorange}
\colorlet{epsilon1color}{tumblue}
\colorlet{epsilon10color}{tumblack}

\colorlet{meadowcolor}{tumbluemedium}
\colorlet{wbarleycolor}{tumbluedark}
\colorlet{corncolor}{tumorange}
\colorlet{wheatcolor}{tumgreen}
\colorlet{sbarleycolor}{tumdiagramred}
\colorlet{clovercolor}{tumdiagramturquoise}
\colorlet{triticalecolor}{tumdiagramsand}

\usepackage[eulergreek]{sansmath}
\pgfplotsset{
	y tick label style={/pgf/number format/.cd,%
		scaled y ticks = false,
		set thousands separator={},
		fixed},
	x tick label style={/pgf/number format/.cd,%
		scaled x ticks = false,
		set decimal separator={,},
		fixed},
	tick label style = {font=\scriptsize},
	every axis label = {
		font=\scriptsize},
	every axis/.append style={
		axis lines=left,
		enlargelimits,
		thick},
	legend style = {font=\scriptsize, draw=none, rounded corners, fill opacity=.5, text opacity=1},
	label style = {font=\scriptsize},
	grid style={line width=.1pt, draw=gray!10},
	major grid style={line width=.2pt,draw=tumgraylight},
}

\pgfdeclarelayer{background}
\pgfdeclarelayer{foreground}
\pgfsetlayers{background,main,foreground}

\author{Marc Ru\ss{}wurm$^1$, Nicolas Courty$^2$, R{\'e}mi Emonet$^3$, S{\'e}bastien Lef{\`e}vre$^2$, Devis Tuia$^1$, and Romain Tavenard$^4$}
\address{$^1$ Environmental Computer Science and Earth Observation Laboratory (ECEO), École Polytechnique Fédérale de Lausanne (EPFL) \\
$^2$ Univ. Bretagne Sud, CNRS, IRISA \\
$^3$Univ Lyon, UJM-Saint-Etienne, CNRS, Institut d Optique Graduate School, Laboratoire Hubert Curien UMR 5516, F-42023 \\
$^4$Univ. Rennes, CNRS, LETG/IRISA\\}
\date{\today}

\journal{ISPRS Journal of Photogrammetry and Remote Sensing}

\title{End-to-End Learned Early Classification of Time Series \\ for In-Season Crop Type Mapping}
\begin{document}

\begin{abstract}
Remote sensing satellites capture the cyclic dynamics of our Planet in regular time intervals recorded in satellite time series data. 
End-to-end trained deep learning models use this time series data to make predictions at a large scale, for instance, to produce up-to-date crop cover maps. 
Most time series classification approaches focus on the accuracy of predictions. However, the earliness of the prediction is also of great importance since coming to an early decision can make a crucial difference in time-sensitive applications.

In this work, we present an End-to-End Learned Early Classification of Time Series (ELECTS) model that estimates a classification score and a probability of whether sufficient data has been observed to come to an early and still accurate decision.
ELECTS is modular: any deep time series classification model can adopt the ELECTS conceptual idea by adding a second prediction head that outputs a probability of stopping the classification.
The ELECTS loss function then optimizes the overall model on a balanced objective of earliness and accuracy.
Our experiments on four crop classification datasets from Europe and Africa show that ELECTS allows reaching state-of-the-art accuracy while reducing the quantity of data massively to be downloaded, stored, and processed.
The source code is available at \texttt{https://github.com/marccoru/elects}.
\end{abstract}

\maketitle

\section{Introduction}
\label{sec:intro}

% research gaps

Efficient large-scale agricultural monitoring and crop type mapping is a prime example of time series analysis in Earth observation: analyzing the temporal variation of vegetation during a growing season is crucial for efficient and accurate predictions.
Models and algorithms trained from satellite time series can distinguish different crop types by observing differences in their respective phenology (life cycles). Traditionally, NDVI-based temporal profiles \cite{wardlow2008large,jonsson2004timesat} are used to extract a fixed set of hand-defined features, such as the date of the green-up, or senescence phases \cite{jonsson2004timesat}.
Remote sensing experts often manually choose the observation period in these approaches to capture the entire vegetative period of the crops in a particular region. The final classification is executed once at the end of this period to produce a crop cover map.
Early time series classification has been a steady topic of interest in remote sensing but is often seen as an auxiliary objective. 
In crop type classification, the terms \emph{in-season-} or \emph{early crop type mapping} are commonly used. 
Several studies \cite{mcnairn2014early,vaudour2015early,inglada2016improved,cai2018high} found that a high classification accuracy for most crop types is achievable within the growing season in a specific region.
A common strategy for assessing which accuracy is possible at what day of the year is \emph{incremental classification}, as termed by Inglada \textit{et al.}, \cite{inglada2016improved}: a supervised classifier performs a classification every time a new image becomes available. The achievable accuracy is then recorded and related to the length of the sequence.
This process involves re-fitting the classifier for different sub-sequences and provides region-specific evidence across all crop types regarding the date at which an accurate classification is possible. 
Recent works have applied incremental classification for early crop type mapping in Germany, \cite{marszalek2022early,kondmann2022early} and South Africa \cite{maponya2020pre}.
Other approaches avoid re-fitting the classifier by choosing sequence-length invariant features \cite{skakun2017early}, employing a cluster-then-labeling strategy \cite{konduri2020mapping}, or modeling simplified two-dimensional feature space in a generative way from historical data \cite{lin2021early}. These approaches employ increasingly sophisticated heuristics to hand-define features invariant to the sequence length. 
Crucially, these approaches yield a rough general knowledge of achievable accuracy given a specific day of the year for a single region across multiple crop types.
Meanwhile, end-to-end deep learning architectures based on recurrence~\cite{russwurm2018multi}, self-attention~\cite{garnot2020psetae}, or convolution~\cite{pelletier2019temporal} can map a variable length series into a fixed-length representation natively.
These deep neural networks learn class-discriminative features solely from a large dataset of labeled samples in an end-to-end scheme by minimizing classification error as the objective function.
To our knowledge, no approach has explicitly optimized a model for the objective of an early classification in remote sensing.

In this work, we address this research gap by End-to-end Learned Early Classification of Time Series (ELECTS). Our method provides early and accurate predictions for each field parcel. To do so, we use a neural network with a loss function optimizing for both objectives: earliness and accuracy.

ELECTS augments and is compatible with recent advances in end-to-end trainable deep time series classification models~\cite{russwurm2018multi,garnot2020psetae,pelletier2019temporal}. As these models produce a fixed-size vector from a variable-length sequence, it does not have to re-fit the classifier on shorter sub-sequences, as earlier incremental classification approaches did \cite{mcnairn2014early,vaudour2015early,inglada2016improved,cai2018high}.
Optimizing on the joint loss objective of earliness and accuracy is also conceptually more straightforward compared to the cluster-then-labeling heuristic of Konduri \emph{et al.}, (2020) \cite{konduri2020mapping} or modeling transitions with two-dimensional distributions from historical data, as Lin \emph{et al.}, (2022) \cite{lin2021early}.

\section{\new{Datasets}}
\label{sec:data}

\begin{figure}
	\includegraphics[width=\linewidth]{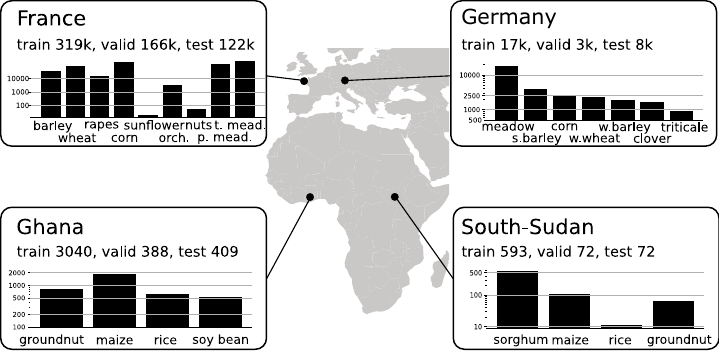}
	\caption{Overview of datasets used in this work. All datasets show a label imbalance, as dominant crops are common in the respective areas. The European datasets BreizhCrops (France) and BavarianCrops (Germany) provide large-scale data with several tens to hundreds of thousands of time series samples. In contrast, the African datasets are smaller and contain a few hundred to a few thousand crop parcels.}
	\label{fig:data}
\end{figure}

\rearrange{

We evaluate ELECTS on four crop-type mapping datasets. Annotations of two datasets originate from crop type statistics collected in Europe and are available at a large scale with several tens of thousand of samples. The annotations of the two datasets in Africa originate from small-scale surveys and contain only hundreds to few thousand annotated time series samples. 
\Cref{fig:data} summarizes the crop-type datasets used in this work. It shows the locations and the label distribution of four crop datasets in Europe and Africa.

\subsection{BreizhCrops (France)}

We use the BreizhCrops dataset \cite{breizhcrops2020} to compare the LSTM model of \cref{sec:model} with several other regular classification models. BreizhCrops contains time series of \num{608263} field parcels of the year $2017$ in Brittany, France.
The time series contains all Sentinel-$2$ images from January to December. Both datasets typically contain between $71$ (every $5$ days) and $147$ (every $2.5$ days) Sentinel-$2$ observations. The high acquisition frequency of $2.5$ days and $147$ observations is possible for some fields in the overlap area of two acquisition stripes.
The BreizhCrops dataset \cite{breizhcrops2020} is split regionally into training (FRH01, FRH02; \num{319258} fields), validation (FRH03; \num{166391} fields), and test (FRH04; \num{122614}) partitions, where FRH$\{1,2,3,4\}$ refers to NUTS-3 administrative boundaries. The \textsl{Nomenclature des unités territoriales statistiques} (NUTS) system delineates Europe in administrative boundaries at three levels: country, state, and province. BreizhCrops uses the division at the provincial level NUTS-3.
The dataset contains nine crop classes: \class{barley}, \class{wheat}, \class{rapeseed}, \class{corn}, \class{sunflower}, \class{orchards}, \class{nuts}, \class{permanent meadows}, \class{temporary meadows}. They are selected to contain both frequent (\class{barley}, \class{wheat}) and rare classes (\class{sunflower}, \class{nuts}), as well as semantically similar categories (\class{permanent-} and \class{temporary meadows}).

\subsection{BavarianCrops (Germany)}

We performed ablation studies and the comparison to one method from the early time-series community (SR2-CF2\cite{mori2017early} in \cref{appendix:comparison:mori}) on a crop type dataset near Hollfeld in Bavaria, Germany, which is a subset of the dataset used in \cite{russwurm2020self}. We chose to subset the original dataset for computational reasons in the initial development and to compare it with existing early time series classification approaches that are typically not designed for large-scale datasets.
Our subset of BavarianCrops covers a $40\text{km} \times 35\text{km}$  area and contains \num{27470} fields that are split into training (\num{16600}), validation (\num{3057}), and test (\num{7813}) partitions, each one organised in blocks of $4.5\text{km} \times 4.5\text{km}$ with \num{500} meter margin between the blocks. All parcels within one block are assigned to the same train-val-test partition to avoid assigning neighboring fields to different partitions \cite{karasiak2021spatial}. Sentinel-$2$ scenes with same frequency as Breizhcrops alongside associated labels are from January to December $2018$ and cover the $7$ common crops \class{meadow}, \class{summer barley}, \class{corn}, \class{winter wheat}, \class{winter barley}, \class{clover}, \class{triticale}. 

\subsection{Ghana and South Sudan}

Rustowicz \emph{et al.} \cite{rustowicz2019semantic} compiled the datasets of Ghana, and South Sudan that were incorporated in the SustainBench dataset \cite{yeh2021sustainbench}. They share a common processing history and are described together in this section.
In these middle- and low-income countries, a substantial portion of the population directly depends on agriculture. An early estimate of the expected crop yield is crucial to evaluate the economic markets and uncover potential shortages. 
This dataset provides Sentinel-2, Sentinel-1, and PlanetScope images of size 64 by 64 pixels from the years 2016 and 2017 linearly interpolated to a time series of 365 days.
We take the imagery and field boundaries and average all pixels belonging to each field to obtain a time series.
Following \cite{rustowicz2019semantic}, the ten Sentinel-2 bands (10m and 20m channels) and NDVI and green chlorophyll vegetation index (GCVI) features are combined with three Sentinel-1 bands (VV, VH, and their ratio) and four PlanetScope bands (RGB+NIR), which results in a 19-dimensional feature vector for each field parcel. While the training and validation datasets were taken from 2016, the samples of the test dataset were taken from the subsequent year 2017.
The crop types classified in Ghana are \class{groundnut}, \class{maize}, \class{rice}, and \class{soy bean}, while information on \class{sorghum}, \class{maize}, \class{rice}, and \class{groundnut} are available in South Sudan.
}

\section{Methodology}
\label{sec:method}

\newcommand{\xuptot}{\M{X}_{\rightarrow t}}
\newcommand{\ptoffset}{\ensuremath{\varepsilon}}

This section describes the details of the proposed method. It consists of a deep learning feature extractor with two decision heads, detailed in \cref{sec:model} and a loss function that optimizes for the dual objective of accuracy and earliness outlined in \cref{sec:loss}.
Throughout this section, we denote vectors with bold-faced symbols, while matrices are bold-faced and capitalized. In a time series, we use $t$ to indicate any time step, while $T$ refers specifically to the index of the last time step in the sequence, i.e., the sequence length.
\Cref{fig:method} shows a schematic view of the model and loss functions with the associated equations of this section.

\subsection{Model}
\label{sec:model}
We use a time series classification model that consists of i) a deep feature extractor based on recursion, $f_\theta$, that ingests time series data one observation at a time and ii) two output heads. This model can be implemented with different deep learning architectures, but we focus on recurrent neural networks (RNNs) without loss of generality. RNNs estimate a hidden representation $\V{h}_t$ at a given time $t$ from an input time series $\xuptot = (\V{x}_1, \V{x}_2, \dots, \V{x}_t)$ of observations $\V{x}$ up to the image acquisition at time $t$.
The model can process a variable number of samples and ingest time series with different sequence lengths $T$.
A recurrent neural network
\begin{align}
\label{eq:feat}
	\V{h}_t = f_{\theta_h}(\V{x}_t, \V{h}_{t-1})
\end{align}
updates its zero-initialized hidden representation $\V{h}_{t-1}$ to $\V{h}_{t}$ with each new observation $\V{x}_t$. It is a natural choice as a feature extractor, as it projects a variable-length input sequence to a fixed-size representation. 
In practice, to avoid vanishing gradients \cite{hochreiter1998vanishing,bengio1994learning}, we choose a Long Short-Term Memory (LSTM) \cite{hochreiter1997long} recurrent neural network $\{\V{h}_t, \V{c}_t\} = f_{\theta_h}(\V{x}_t, \{\V{h}_{t-1}, \V{c}_{t-1}\})$ that updates two hidden representations where we use the cell output $\V{h}_t$ for two linear decision heads: one head produces a classification probability for each class 
\begin{align}
	\label{eq:dec}
	\hat{\V{y}}_t = \text{softmax}\left(f_{\theta_c}(\V{h}_t)\right)% = \text{softmax}\left(\V{h}_t\theta_c\right)
\end{align}
and another one outputs a scalar probability of stopping
\begin{align}
	\label{eq:class}
	d_t = \sigma\left(f_{\theta_d}(\V{h}_t)\right)% =  \V{h}_t\theta_d
\end{align}
the classification decision. The $\sigma$ symbol denotes the sigmoid function that rescales the outputs of the stopping head to a probability between 0 and 1. At test time, a hard stopping decision is sampled
from this stopping probability. As an example: with a stopping probability $d_t = 0.2$, the classification is stopped with a 20\% probability at this time. In practice (see \cref{fig:introdatafig} in results), we observe that $d_t$ raises sharply from 0 to 1 within a few time points on a trained model. 

\begin{figure*}[!t]
	\centering\tikzsetnextfilename{methods_figure}

\colorlet{modelcolor}{acier!10}
\colorlet{losscolor}{rouge!5}

\tikzstyle{rnn}=[draw,rounded corners]
\tikzstyle{annot}=[rounded corners, fill=colorblue!20]
\tikzstyle{infer}=[-stealth, shorten >=.0em, shorten <=.0em, colorinfer]
%\tikzstyle{loss}=[rounded corners, font=\small]
\tikzstyle{grad}=[colortrain]
\begin{tikzpicture}[node distance=1em]
\node[](x0){$\V{x}_t$};
\node[rnn, below=2em of x0](f0){$f_{\theta_h}(\V{x}_t, \V{h}_{t-1})$};  %$\theta_\text{rnn}$};
\node[below= of f0](h0){$\V{h}_t$};  %$\theta_\text{rnn}$};
\node[right=0em of f0, font=\tiny, yshift=.5em]{\cref{eq:feat}};

\node[rnn, right=of h0](fy0){$f_{\theta_c}(\V{h}_t)$};
\node[below=of fy0](y0){$\hat{\V{y}}_t$};

\node[rnn, left=of h0](fd0){$f_{\theta_d}(\V{h}_t)$};
\node[below=of fd0](d0){$\hat{d}_t$};

\node[above=-.2em of fd0, font=\tiny, xshift=-.5em]{\cref{eq:class}};
\node[above=-.2em of fy0, font=\tiny, xshift=.5em]{\cref{eq:dec}};

\draw[infer] (f0) -- (h0);
\draw[infer] (x0) -- (f0);

\draw[infer] (x0) -- (f0);
\draw[infer] (h0) -- (fy0);
\draw[infer] (h0) -- (fd0);
\draw[infer] (fy0) -- (y0);
\draw[infer] (fd0) -- (d0);

%	\node[above=0.2em of f0,font=\scriptsize, xshift=3em](modellabel){\cref{sec:model}};

\begin{pgfonlayer}{background}
	\node[fit=(fd0)(fy0)(f0), draw=acier, rounded corners, fill=modelcolor] (model) {};
\end{pgfonlayer}

{
\node[left=5em of f0](hprev){$\V{h}_{t-1}$};
\draw[infer] (hprev) -- (f0);
\node[above=2em of hprev](xprev){$\V{x}_{t-1}$};
\draw[infer] (xprev) -- (hprev);
\node[below=of hprev, yshift=-.25em] (phantomhtprev) {\phantom{$\V{h}_t$}};
\begin{pgfonlayer}{background}
	\node[fit=(hprev)(phantomhtprev), draw=acier, rounded corners, fill=modelcolor] (prevmodel) {};
\end{pgfonlayer}
%	\node[below=of phantomhtprev, xshift=1.5em](yprev){$\hat{\V{y}}_{t-1}$};
\node[below=of phantomhtprev](dprev){$\hat{d}_{t-1}$};
%	\node[loss, below=of yprev](Lcerprev){$L_{\text{CER}}(\hat{\V{y}}_{t-1}, \V{y})$};
\draw[-stealth] (phantomhtprev) -- (dprev);
}	

{
\node[left=2em of hprev](hprevprev){$\V{h}_{t-2}$};
\draw[infer] (hprevprev) -- (hprev);
\node[above=2em of hprevprev](xprevprev){$\V{x}_{t-2}$};
\draw[infer] (xprevprev) -- (hprevprev);
\node[below=of hprevprev, yshift=-.25em] (phantomhtprevprev) {\phantom{$\V{h}_t$}};
\begin{pgfonlayer}{background}
\node[fit=(hprevprev)(phantomhtprevprev), draw=acier, rounded corners, fill=modelcolor] (prevprevmodel) {};
\end{pgfonlayer}
%	\node[below=of phantomhtprev, xshift=1.5em](yprev){$\hat{\V{y}}_{t-1}$};
\node[below=of phantomhtprevprev](dprevprev){$\hat{d}_{t-2}$};
%	\node[loss, below=of yprev](Lcerprev){$L_{\text{CER}}(\hat{\V{y}}_{t-1}, \V{y})$};
\draw[-stealth] (phantomhtprevprev) -- (dprevprev);
}	
%	\node[left=of prevprevmodel]{\dots};
\node[left=of dprevprev](dprevprevprev){\dots};
\node[left=of xprevprev]{\dots};

%%%%% NEXT
\node[right=4em of f0](hnext){\dots};
\draw[infer,draw=lightgray] (f0) -- (hnext);

%%%% NEXTNEXT
\node[right=of hnext](hnextnext){$\V{h}_{T}$};
\draw[infer] (hnext) -- (hnextnext);
\node[above=2em of hnextnext](xnextnext){$\V{x}_{T}$};
\draw[infer] (xnextnext) -- (hnextnext);
\node[below=of hnextnext, yshift=-.25em] (phantomhtnextnext) {\phantom{$\V{h}_t$}};
\begin{pgfonlayer}{background}
\node[fit=(hnextnext)(phantomhtnextnext), draw=acier, rounded corners, fill=modelcolor] (nextnextmodel) {};
\end{pgfonlayer}
\node[below=of phantomhtnextnext](dnextnext){$\hat{d}_{T} = 1$};

%%%% LOSSES

\node[below=of y0](L0){$L_{\text{CER}}(\hat{\V{y}}_{t}, \V{y})$};
\draw[-stealth] (y0) -- (L0);

\node[below=-.2em of L0, font=\tiny, xshift=1.5em]{\cref{eq:lcer}};

\node[below=of d0, yshift=.2em](D) {$D(\hat{\V{d}}_{\rightarrow t})$};

\draw[infer, rounded corners](dprev) |- ($(D.west)+(0,.1)$);
\draw[infer, rounded corners](dprevprev) |- ($(D.west)+(0,0)$);
\draw[infer, rounded corners, draw=lightgray](dprevprevprev) |- ($(D.west)+(0,-.1)$);
\draw[infer](d0) -- (D);

\node[below=-.3em of D, font=\tiny, xshift=1.5em]{\cref{eq:d}};

\node[below=of L0, xshift=-2em] (Lelects) {$L_{\text{ELECTS}}^t(\hat{\V{d}}_{\rightarrow t},\hat{\V{y}}_t,\V{y})$};
\node[below=-.6em of Lelects, font=\tiny, xshift=3.5em](labellelects){\cref{eq:lelects}};
\draw[infer] (D) |- (Lelects);
\draw[infer] (L0) -- ++(0,-2em);

\begin{pgfonlayer}{background}
	\node[fit=(Lelects)(D)(L0), draw=rouge, rounded corners, fill=losscolor] (loss) {};
\end{pgfonlayer}

%	\node[above=0.1em of loss,font=\scriptsize, xshift=0em](modellabel){\cref{sec:loss}};

\node[below=0mm of xprev |- Lelects.north, fill=losscolor, draw=rouge, rounded corners](Lprev) {$L^{t-1}$};
\node[below=0mm of xprevprev |- Lelects.north, fill=losscolor, draw=rouge, rounded corners](Lprevprev) {$L^{t-2}$};
\node[below=0mm of dprevprevprev |- Lelects.north, rounded corners](Lprevprevprev) {\dots};
\node[below=0mm of xnextnext |- Lelects.north, fill=losscolor, draw=rouge, rounded corners](Lnextnext) {$L^{T}$};

\node[below=of loss](sumloss){$\sum_{t=1}^{T} L^{t}_{\text{ELECTS}}$};
\node[below=-0.5em of sumloss,font=\scriptsize, xshift=3em]{\cref{eq:argmin}};

\draw[infer, rounded corners, draw=gray](Lprevprevprev) |- ($(sumloss.west)+(0,-.1)$);
\draw[infer, rounded corners](Lprevprev) |- ($(sumloss.west)+(0,0)$);
\draw[infer, rounded corners](Lprev) |- ($(sumloss.west)+(0,.1)$);
\draw[infer, rounded corners](Lnextnext) |- ($(sumloss.east)$);
\draw[infer, rounded corners](Lelects) -- ++(0, -2em);

\node[left=of prevprevmodel, font=, text width=2cm](modellabeldesc){Model \\ (\cref{sec:model})};
\node[below=0mm of modellabeldesc |- Lprevprev.north, font=, text width=2cm, yshift=1em]{Losses \\ (\cref{sec:loss})};

%	\node[below=of d0](pt){$p(t_\text{dec}=t | \xuptot)$};
%	\node[below=8em of f0, loss](L){$\mathcal{L}(\xuptot, y ; \alpha)$}; % = P(t)\mathcal{L}_c (\xuptot, y)$};
%	
%	\node[left=1em of pt](budget){$p(t_\text{dec} < t | \xuptot)$};

%	\draw[infer,dashed] (L0) -- (L);
%	\draw[infer,dashed] (d0) -- (pt);
%	\draw[infer,dashed] (pt) -- (L);
%	\draw[infer,dashed] (budget) -- (pt);

\end{tikzpicture}
	\caption{Schematic illustration of model (blue) and losses (red) of this \cref{sec:method}. Arrows indicate function inputs and outputs. Neural network components are denoted by $f_\theta$. At test time, with fixed weights, $\theta$, the model (blue) can process a time series up to any time $t$. At training time, losses are calculated on the complete time series until the last time step $T$.
	}
	\label{fig:method}
\end{figure*}

\subsection{ELECTS loss function}
\label{sec:loss}

At each time $t \leq T$, we compute the classification, earliness-rewarded loss, $L_{\text{CER}}$:

\begin{align}
\label{eq:lcer}
L_{\text{CER}}(\hat{\V{y}}_t, \V{y}) = \alpha L_c(\hat{\V{y}}_t, \V{y}) - (1-\alpha)R_e(\hat{\V{y}}_t, \V{y}, t).
\end{align}

We weight both terms with an $\alpha \in [0,1]$ hyper-parameter that trades off accuracy and earliness reward.
The classification loss is the negative log-likelihood or cross-entropy loss
\begin{align}
	L_c(\hat{\V{y}}_t, \V{y}) = - \sum_{c=1}^{C}y_c\log{\hat{y}_{c,t}},
\end{align}
while the earliness reward is
\begin{align}
	R_e(\hat{\V{y}}_t,\V{y}, t) = \hat{y}_t^+\left(\frac{T-t}{T}\right).
\end{align}

As such, $R_e$ decreases linearly for later predictions when $t$ approaches $T$. This term is scaled with the probability of the correct class $y_t^+ = \sum_{c=1}^{C}y_c\hat{y}_{c,t}$ with $\V{y}$ as one-hot vector of C classes. 
This term applies the reward only if the probability of the correct class is large.

$L_{\text{ELECTS}}$ is computed for each time $t$ in a training sample time series of length $T$ to minimize a joint expression of accuracy (via $L_\text{CER}$) and explicit earliness (via $D_t$) as:
\begin{align}
	\label{eq:lelects}
	L_{\text{ELECTS}}(\hat{\V{d}}_{\rightarrow t},\hat{\V{y}}_t,\V{y}) = {D}_t(\hat{\V{d}}_{\rightarrow t}) L_{\text{CER}}(\hat{\V{y}}_{t}, \V{y})
\end{align}
where
\begin{align}
	\label{eq:d}
	D(\hat{\V{d}}_{\rightarrow t}) = \hat{d}_t \prod_{i=1}^{t-1}(1-\hat{d}_i) + \frac{\varepsilon}{T}
\end{align}
can be interpreted as the joint probability of making a decision $\hat{d}_t$ at time $t$ and not having made a decision before $\prod_{i=1}^{t-1}(1-\hat{d}_i)$.
At the last time step, we set $\hat{d}_T = 1$ irrespectively of the model output to make sure
that the model has taken a stopping decision in the interval $[0, T]$. In practice, we add a small constant offset $\frac{\varepsilon}{T}$ to each $\hat{d}_t$, with $\varepsilon$ as an hyper-parameter.
This offset makes $D(\hat{\V{d}}_{\rightarrow t})$ non-zero for all $t$, which encourages the model to make accurate classifications for all time steps in \cref{eq:lelects}. With $\varepsilon=0$, only accurate classifications at the time steps close to the stopping time, where $D(\hat{\V{d}}_{\rightarrow t})$ is large, would be encouraged.
Without this offset, i.e., with $\varepsilon=0$, we found experimentally (see \cref{appendix:ablations}) 
that a randomly-initialized model tended to fall in a local minimum when optimizing \cref{eq:lelects} by predicting early at low accuracy.

The learnable parameters $\theta_h, \theta_d, \theta_c$ are determined by minimizing the overall objective
\begin{align}
	\label{eq:argmin}
	 \argmin_{\theta_h, \theta_d, \theta_c} \sum_{\M{X},\V{y}}\sum_{t=1}^T L_\text{ELECTS}(\underbrace{f(\M{X}; \theta_h, \theta_d, \theta_c)}_{\hat{\V{d}}_{\rightarrow t},\hat{\V{y}}_t}, \V{y})
\end{align}
for each time $t$ over a dataset of labeled samples $\M{X}, \V{y}$.

\rearrange{
\subsection{\new{Implementation Details}}
\label{sec:implementationdetails}

For all results described in the \cref{sec:results}, we used a recurrent neural network with the same hyperparameters for all datasets.
An initial linear layer (and layer normalization) projects the original input vector to a learned $32$-dimensional feature representation at each time, followed by two mono-directional LSTM layers.
We implement each decision head as a linear layer with a sigmoid activation function for the stopping decision and softmax for the classification scores, respectively. The overall model has \num{67108} trainable parameters, making this implementation light weighted and trainable in any desktop machine with a GPU graphics card. As stated above, researchers can implement the ELECTS loss on any neural network for time series making it adaptable for other time series approaches.
We used the Adam optimizer with a learning rate of $0.001$ and a dropout of $20$\%. We determined these hyperparameters experimentally on the validation set of the BavarianCrops dataset (described in the next section). With a batch size of $256$, we trained models in a few minutes (BavarianCrops) or a few hours (BreizhCrops) on a  GeForce RTX $3090$. For BavarianCrops and BreizhCrops, we randomly choose sequences of $70$ observations from the originally longer complete time series to obtain sequences of equal length for training in batches.
At test time, we can run inference on the complete variable-length time series. 
For the Ghana and South Sudan datasets, we train on the interpolated 365-day sequences, similar to \cite{rustowicz2019semantic}.
We used an $\varepsilon=10$ offset parameter throughout the experiments with a fixed sequence length of the respective training dataset.
}

\subsection{Model Evaluation}

We evaluate the model on the four different crop type mapping datasets in Europe and Africa, described in \cref{sec:data}.
We train, validate, and evaluate the model for each dataset on spatially disjoint training, validation, and test regions. For BavarianCrops, training and evaluation fields were separated by blocks, while different administrative boundaries were used in BreizhCrops. For Ghana and South Sudan, we followed the split of Rustowicz \emph{et al.}, (2019) \cite{rustowicz2019semantic}.
For each dataset, we re-train the ELECTS-LSTM model from scratch on the respective training dataset and evaluate the performance on the test set.
We do not vary the hyper-parameters (network layers, hidden dimensions, learning rates) across the datasets in these experiments and keep the identical model architecture throughout this work.

\section{Results}
\label{sec:results}

This section presents the results obtained with the ELECTS-trained LSTM neural network described in \cref{sec:method}. 
We structure this section in three parts: First, \cref{sec:accuracy} shows the prediction process on individual field parcels qualitatively and quantitatively. \Cref{sec:stoppingtimes} focuses on the dates of stopped decisions and relates these to phenological events. It provides interpretations of the model predictions on two crop classes (\class{rapeseed} and \class{barley}). 
\new{For these experiments, we used the large BreizhCrops dataset in \cref{sec:accuracy,sec:stoppingtimes}.}
Finally, we expand the scope in \cref{sec:applicability}, where we train the ELECTS recurrent neural network on multiple datasets in Europe \new{(BreizhCrops, BavarianCrops)} and Africa \new{(Ghana, South Sudan)}, as outlined in \cref{sec:data}.
Further model comparisons developed in the time series community and ablations on the loss design on the BavarianCrops dataset are reported in \cref{appendix:comparison,appendix:ablations}, respectively.

\subsection{Accuracy Evaluation}
\label{sec:accuracy}

\cref{sec:singlefieldprediction,sec:maps} illustrate the prediction process, while \cref{sec:quantitativeprediction}, analyzes the classification accuracy of the stopped fields quantitatively throughout the year on all field parcels in the BreizhCrops test set.

\subsubsection{Single Field Prediction}
\label{sec:singlefieldprediction}

    \begin{figure*}
        \centering\includegraphics[width=.75\linewidth]{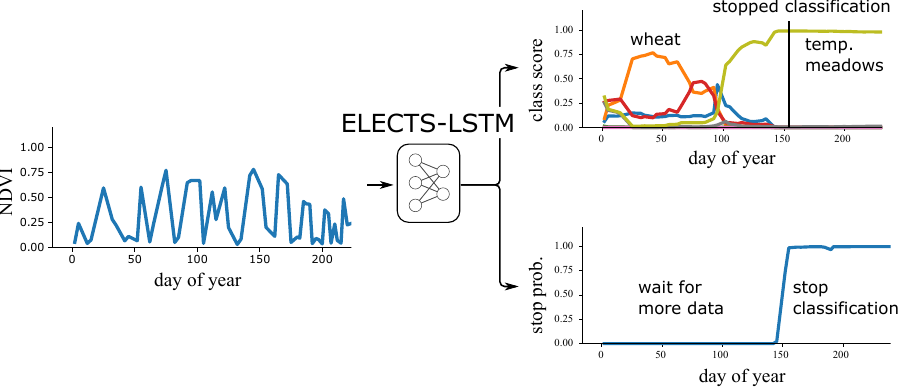}
        \caption{Prediction of the ELECTS-trained early classification model. The model ingests a time series (left) incrementally one element at a time. It estimates a probability for each crop category (top right) alongside a probability of stopping (bottom right). As long as the probability of stopping remains low, more data is necessary to obtain an accurate classification result.}
        \label{fig:introdatafig}
    \end{figure*}

\Cref{fig:introdatafig} illustrates the prediction process with the ELECTS-trained LSTM on a single time series sample from the BreizhCrops test set. 
The time series of this \class{temporary meadow} field is represented on the left as its NDVI profile. 
\new{However, note that our model uses thirteen spectral bands' complete signal at each observation.} 
This profile shows that this field parcel is photosynthetically active (high NDVI) across the year. These high-NDVI observations in this time series are interrupted by negative outliers caused by cloud cover (low NDVI). 
The ELECTS-trained LSTM neural network ingests this time series \new{one-time} step at a time and estimates a probability for each crop class (top right) and a probability of stopping (bottom right). The model estimates a high probability for the class \class{wheat} (orange) during the first hundred days of the year. If the model stopped the classification decision this early, it would incorrectly predict the class \class{wheat}. Further, during the first hundred days, the probability of stopping remains low, indicating that more data is necessary for a confident decision. From the day of year 100 onwards, the model assigns the highest classification probability to the correct class \class{temporary meadow}.
The probability of stopping remains low until day-of-year 150 when a rapid increase indicates that the model is sufficiently confident to stop the classification.

\subsubsection{Classification of Fields at Different Times}
\label{sec:maps}

\begin{figure*}
	\centering\input{images/maps.tikz}
	\caption{\new{The ELECTS prediction process is shown} in a deployment setting for one year, where the model predicts class labels and stops the predictions of individual fields. It shows early classification results in a $2.5\text{km}~\times~2.5\text{km}$ site in the Brittany test region (\num{2.4052}$^\circ$  West, \num{47.5328}$^\circ$  North). The final results are shown in the right column. The previous columns show three dates with associated predictions (second row), and a correct/incorrect map (third row, with blue being correct predictions and red incorrect). Transparency is used for predictions in these rows to de-emphasize the fields where the prediction has stopped.
		The fourth row shows a binary score specifically to indicate which classifications are still active (white) or have already been stopped (black). The bottom right image depicts the stopping day for all parcels.}
	\label{fig:maps}
\end{figure*}

\Cref{fig:maps} shows a crop cover map of $250$ field parcels from a $2.5\text{km}~\times~2.5\text{km}$ area of interest within Brittany, France, from the BreizhCrops test set.
It illustrates the prediction process in the deployment setting where the predictions of some fields are stopped at different times compared to others in the same geographic area. 
The top row presents RGB images from this area alongside the ground truth crop type.
The other rows show the model predictions at each date (second row), the correct/incorrect predictions (third row, with blue being correct and red incorrect), and the active (white) vs. stopped (black) status of each parcel. 
The rightmost column shows the predictions after recombining all predictions obtained at the respective stopping time (second and third row) and a summary of the per parcel stopping date. 
For ELECTS, only a stopped field (black) classification decision is relevant, as active fields (shown in white) require more data. 
In rows two and three's prediction and correctness figures, we present parcels still active in the decision process with transparent colors. The field parcels where the model stopped the classification process are drawn with opaque colors without transparency.

On April $12$th (first column in \cref{fig:maps}), most parcels were covered homogeneously with green vegetation. The model predicted most fields as \class{temporary meadow} and \class{corn}, among the dataset's most frequent classes. The overall accuracy for these parcels was $54$\%. These early, incorrect classifications (red) are frequent, as not enough time series could be observed this early in the year. The ELECTS-LSTM model did not stop any fields at this point (no black fields in the bottom row). More time steps are required for a confident classification decision.
On May $22$nd (second column in \cref{fig:maps}), the overall accuracy has increased to $74$\%. At this date, several parcels of \class{corn} (in red) and one of \class{rapeseed} (green) were stopped and correctly classified. The model predictions did not vary noticeably in June $21$st (third row in \cref{fig:maps}) \new{concerning April $12$th}. However, the number of stopped parcels increased steadily, as shown in the bottom row.
The model classified most fields within the year's first half (shown in day-of-stopping; last row and column). However, single fields were still classified later in the year, emphasizing the need for a stopping decision for each field parcel, as the ELECTS-LSTM model provides.

\subsubsection{Quantitative Prediction Accuracy}
\label{sec:quantitativeprediction}

\begin{figure}
	\includegraphics[width=\linewidth]{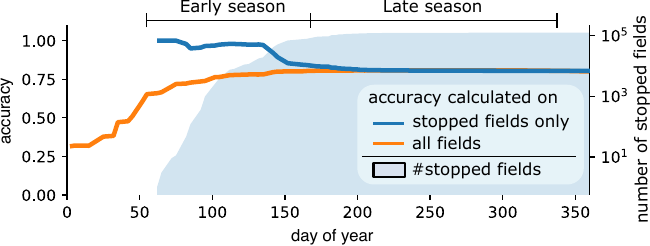}
	\caption{The accuracy of the ELECTS model during the year was calculated on all fields (orange) and only the stopped fields (blue).
		The ELECTS model only stops fields from doy (day-of-year) 60 onwards in the early season. The stopped fields are classified more accurately compared to all fields in the test region. A user can expect an accurate prediction of a stopped field.}
	\label{fig:confcomparison}
\end{figure}

\Cref{fig:confcomparison} shows the classification accuracy up to a specific day of the year for all field parcels (orange) and only the stopped field parcels (blue) in the test set of BreizhCrops in Brittany, France.
The horizontal axis represents the prediction date within the year until data is accessible to the classifier. The blue shaded area shows the number of stopped fields at each respective date.
The orange line shows the classification accuracy calculated on all fields and represents the performance of a regular accuracy-only time series model.
The blue line shows the classification accuracy only of the stopped fields, as provided by an early classification approach, such as the ELECTS-LSTM model in this work.

Early in the year, only a low accuracy between 30\% and 50\% on all fields is possible before March $1$st, the day-of-year (doy) 60. The model can observe no fine-grained classification-relevant features this early, as this period falls in the winter season. The stopping decision of the early classification model reflects this by declaring no fields as stopped before doy 60.
From doy 60 onwards, an increasing number of fields are declared as stopped, as indicated by the blue shaded area.
Notably, the few fields the model stopped in the early season between March and June (doys 60 and 150) are predicted at high accuracy, as shown by the blue line. 
Later during the year, the accuracy decreases when the classification of the majority of fields is stopped.
The high accuracy of early-stopped fields reflects the intuition that the stopping decision is related to the model's confidence, as presumably easier-to-classify fields are stopped first, leading to the high accuracy in the early season. 
The more ambiguous and difficult fields are stopped in the late season. Wrong classifications become more common, and the accuracy drops to the same level as expected by an accuracy-only classifier. 

From a practical deployment perspective, this result demonstrates that a user can be confident that the predictions are accurate for the stopped fields. This allows the user to make decisions for these individual field parcels early in the season. 

\subsection{\new{Earliness Evaluation}}
\label{sec:stoppingtimes}

\begin{figure*}
	
	\begin{subfigure}{.49\textwidth}
		\includegraphics[width=\textwidth]{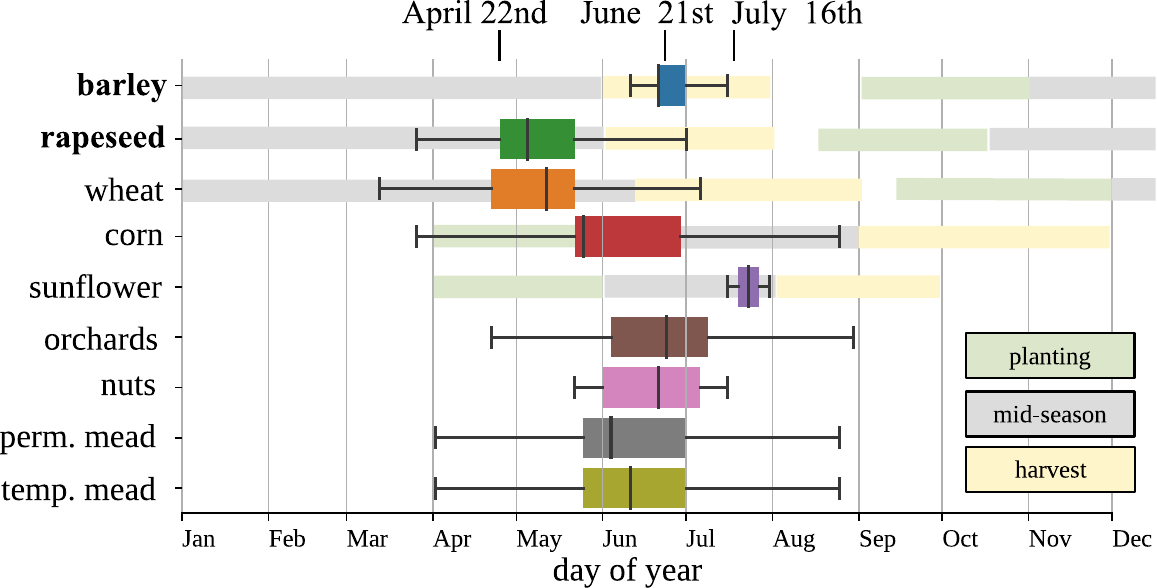}
		\caption{Quantitative evaluation of stopping times per crop type in Brittany, France, overlayed with planting, mid-season, and harvest dates from the crop calendar for France by the USDA Foreign Agricultural Service.}
		\label{fig:boxplots}
	\end{subfigure}
	\begin{subfigure}{.49\textwidth}
            \resizebox{\textwidth}{!}{
		\centering
		\begin{tikzpicture}[node distance=0em]
			\node[label={[rotate=90, anchor=south]left:barley}](a){\includegraphics[width=3cm]{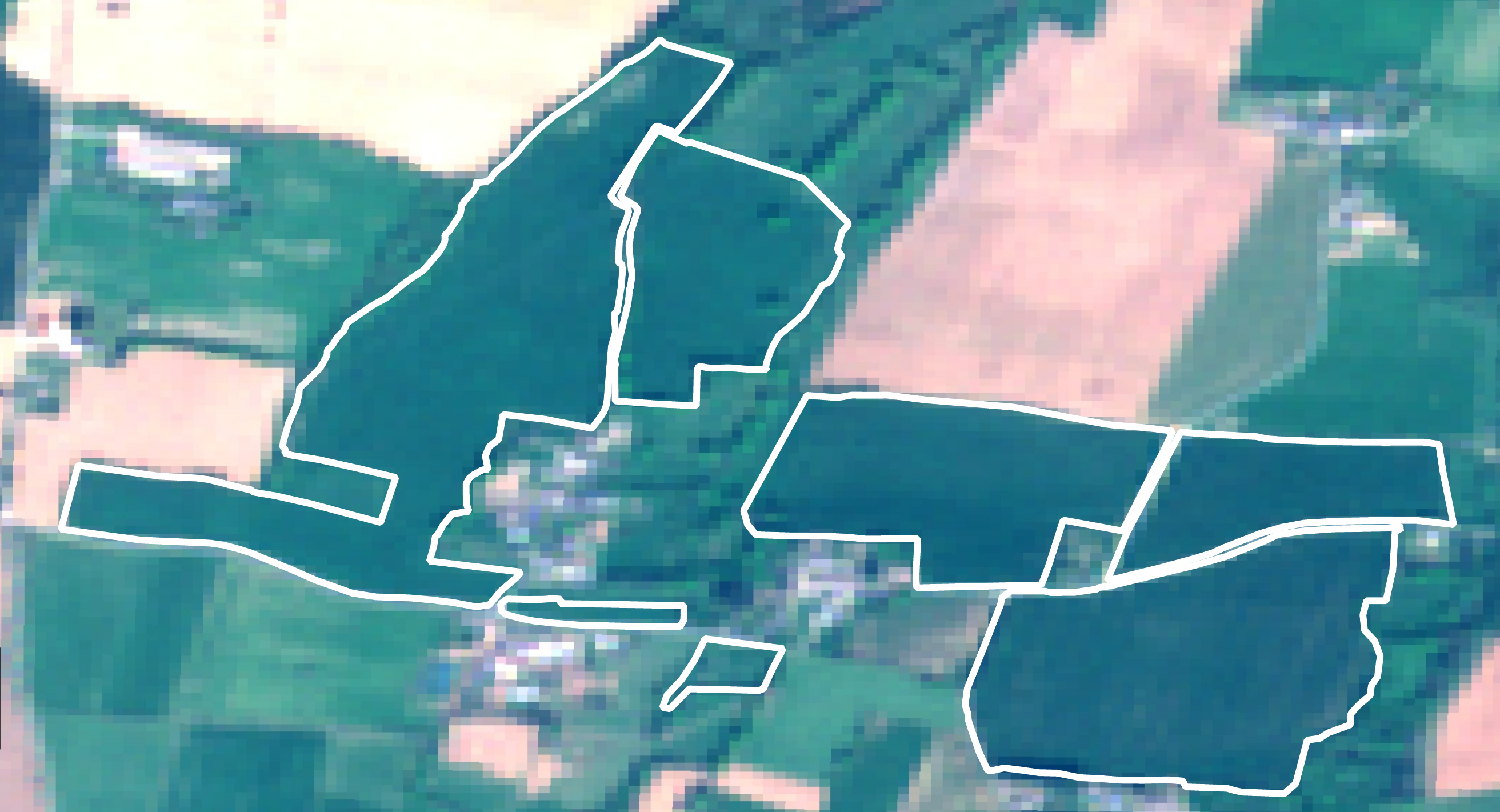}};
			\node[right=of a](b){\includegraphics[width=3cm]{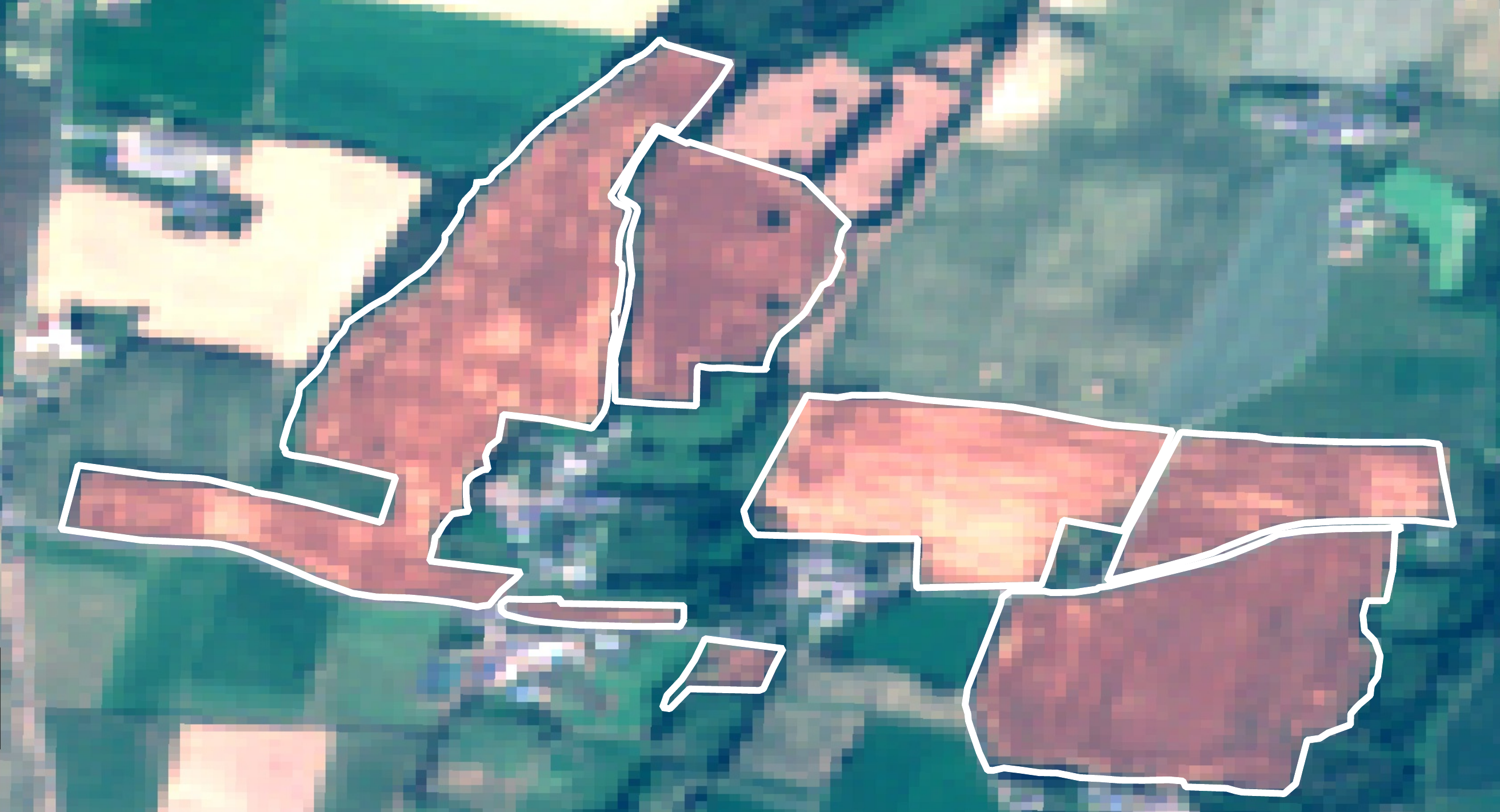}};
			\node[right=of b](c){\includegraphics[width=3cm]{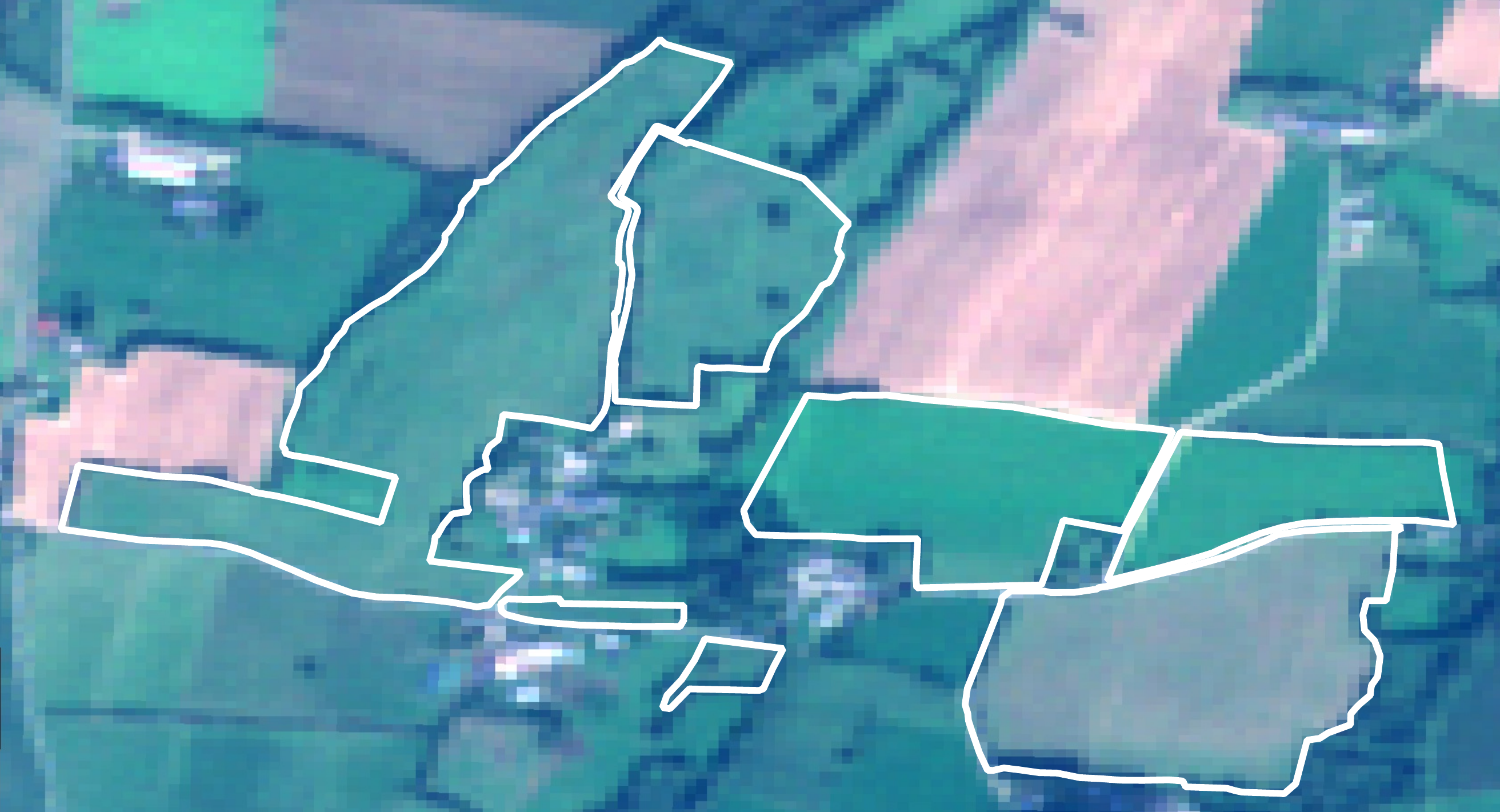}};
			
			\node[label=above:2017-04-22, above=of a, label={[rotate=90, anchor=south]left:rapeseed}]{\includegraphics[width=3cm]{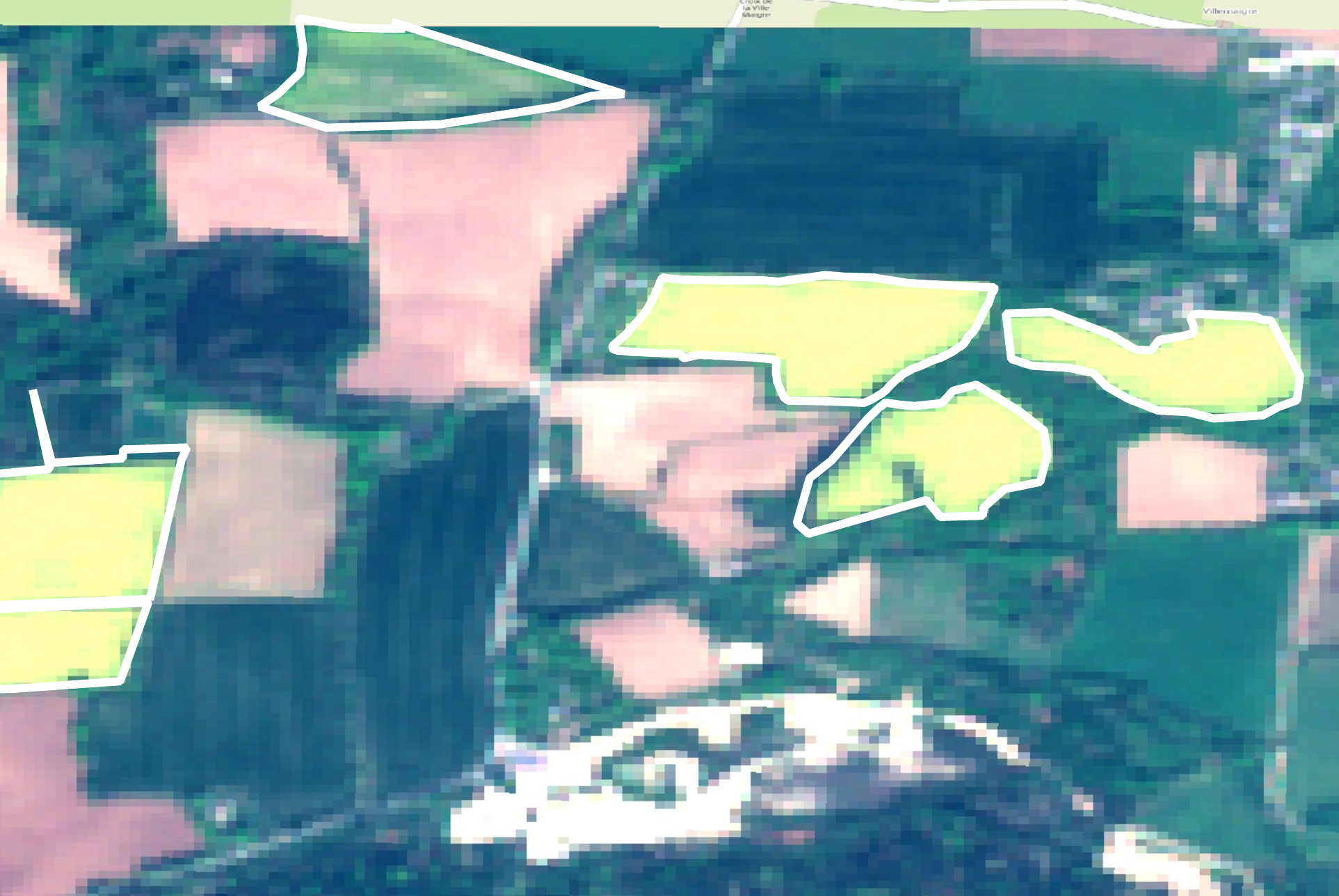}};
			\node[above=of b, label=above:2017-06-21]{\includegraphics[width=3cm]{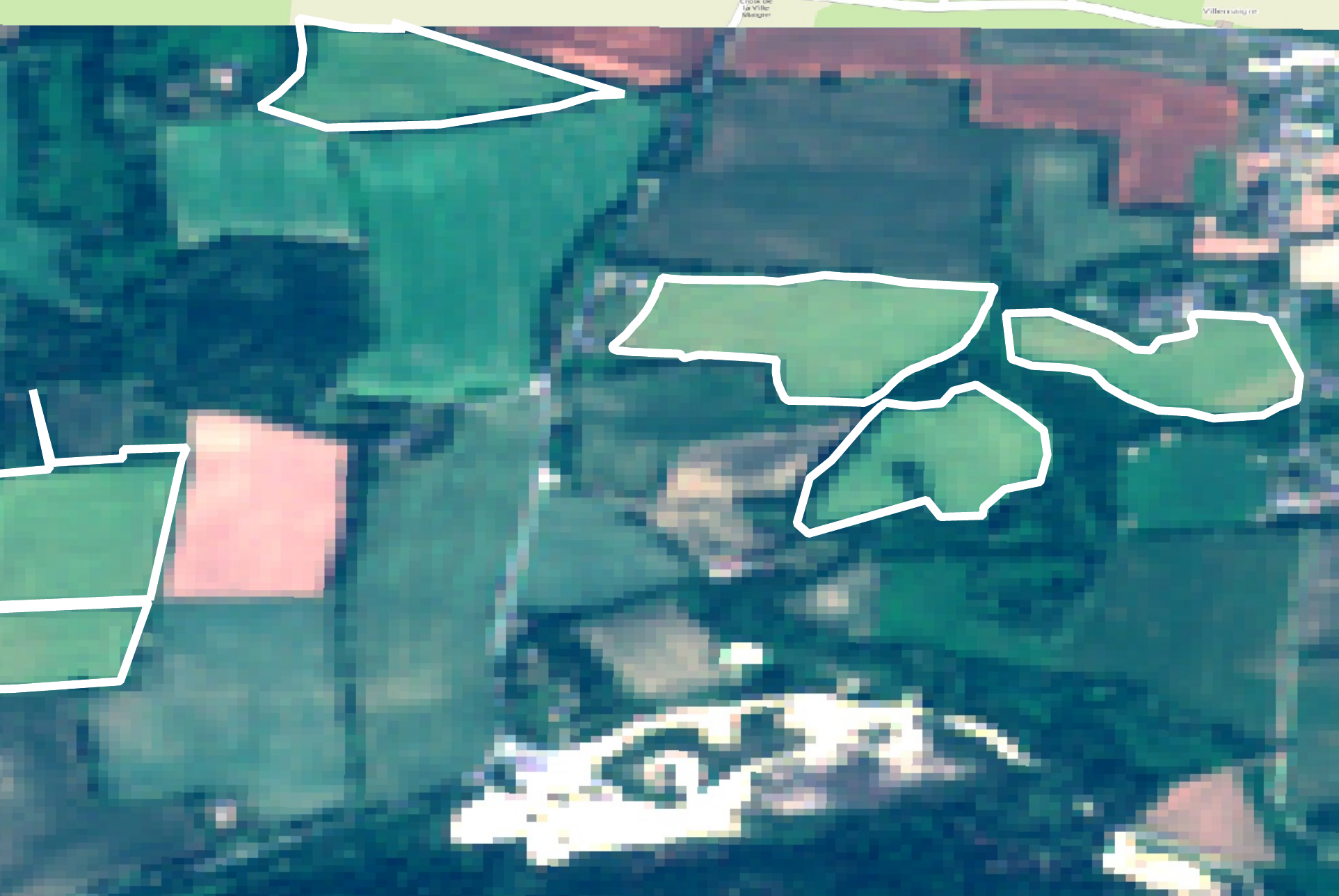}};
			\node[above=of c, label=above:2017-07-16]{\includegraphics[width=3cm]{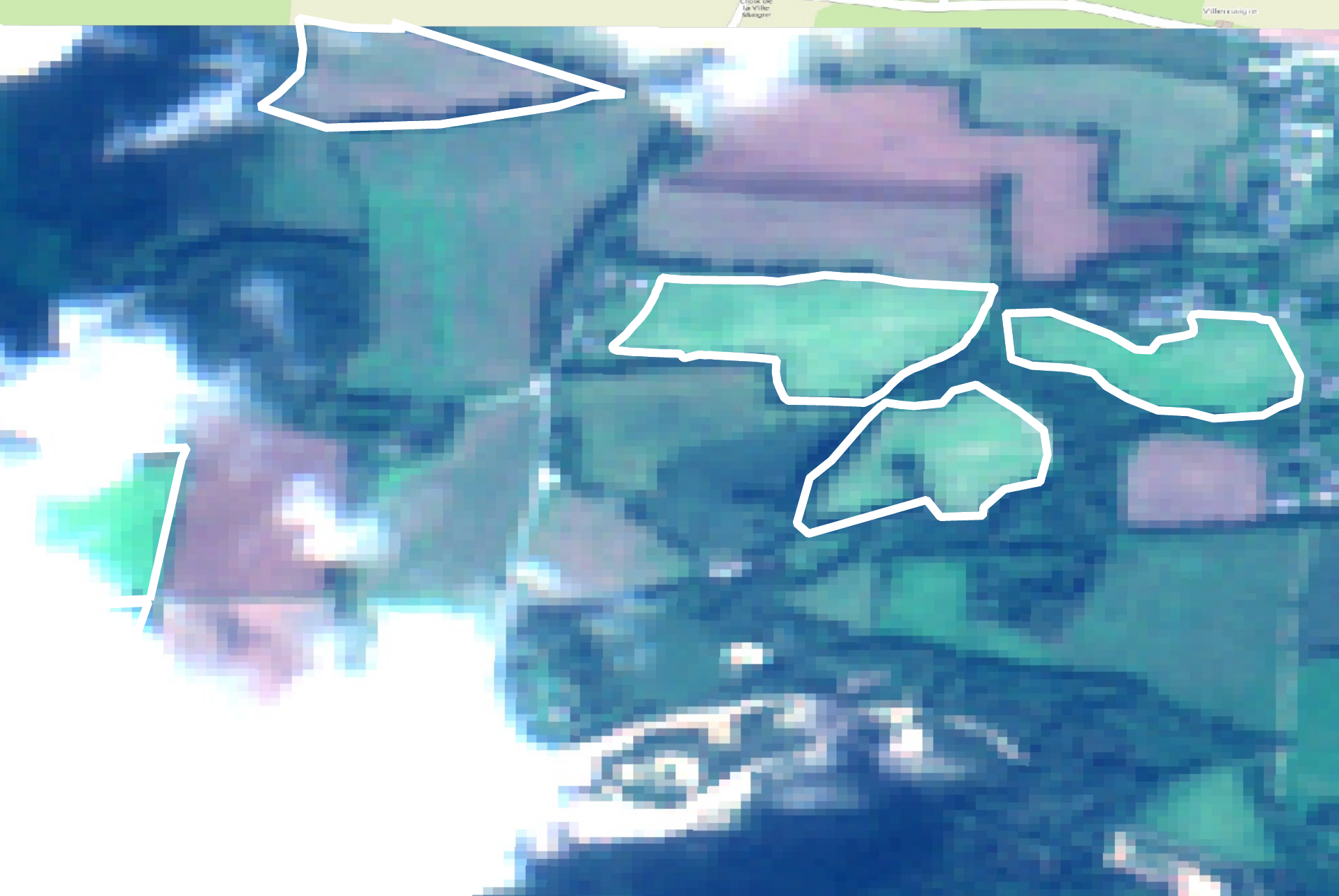}};
		\end{tikzpicture}
            }
		\caption{Images of \class{barley} and \class{rapeseed} fields on three selected dates. The model stopped the classification of \class{rapeseed} fields end of April when the characteristic yellow blossoms were visible, shown in the April $22$nd image. \class{Barley} field classifications were stopped end of June during the harvest, as visible in June $21$st, where bare soil is visible on the field parcels.}
		\label{fig:crop_calendar:qualitative}
	\end{subfigure}
	
	\caption{Stopping times of individual crops classes related to the local crop calendar in Brittany, France and examples of \class{barley} and \class{rapeseed} fields that reveal that specific machining and blossoming events cause early classifications of these crop categories.}
	\label{fig:crop_calendar}
\end{figure*}

\cref{fig:crop_calendar} analyses the dates in which the ELECTS model stopped the classification from a phenological perspective concerning a local crop calendar of France. 
\Cref{fig:boxplots} shows that stopping dates vary for individual crop types where the classification of all crops has been stopped in the agriculturally relevant period between planting and harvest.
The average classification time of most crops, i.e., \class{wheat}, \class{rapeseed}, \class{corn}, \class{sunflower}, lies in the mid-season period.
Notably, all crops except \class{barley} (discussed later) were classified before the harvest period, which domain experts often consider the end-of-series date for accuracy-only classifiers when knowledge of local crop calendars is available.

\class{Rapeseed} parcels were classified particularly early: towards the end of April until mid-May, two months before the harvest period. 
We analyze these crop fields qualitatively in \cref{fig:crop_calendar:qualitative} where several \class{rapeseed} fields are highlighted by a white outline on images from April 22nd, June 21st, and July 16th.
\class{Rapeseed} fields blossom in a characteristic yellow color, as visible in the image of April 22nd. This blossoming period falls into the window \new{where} the classification of the majority of \class{rapeseed} parcels has been stopped. 
Hence, we can \new{deduce} that the model uses this blossoming event as a characteristic feature to classify these parcels as \class{rapeseed} and stop the classification confidently.
In \cref{fig:boxplots}, \class{barley} was the only crop type where the model stopped the prediction during the harvest period end of June. We analyze this period qualitatively in the second row of \cref{fig:crop_calendar:qualitative} that shows \class{barley} parcels outlined in white color.
In particular, the effects of harvest are visible on June 21st, where \class{barley} fields are the only field parcels that were recently harvested. After that date, bare soil is observed in the parcel, while all neighboring field parcels are covered by vegetation. From this analysis, we can deduce that these harvest operations cause the stopping decisions of the \class{barley} crops, as the stopping dates of \class{barley} parcels fall narrowly into this period.
This analysis shows that the stopping times produced by the ELECTS-LSTM early classification model fall into a meaningful phenological period for this region. The interpretation of the crop calendar and explanation of the \class{barley} and \class{rapeseed} parcels show that the model learned to utilize meaningful features (e.g., the blossoming event of \class{rapeseed}) to come to the stopping decision.
Notably, this is learned without any direct temporal supervision as the model is optimized end-to-end solely on crop labels without any labels on time or crop cycles for this area.

\subsection{Applicability \new{of ELECTS} across datasets}
\label{sec:applicability}

The proposed model can be trained end-to-end on any time series classification problem if sufficient class-labeled data is available.
This enables us to train models for different geographic areas without requiring region-specific expert knowledge aside from the labeled samples in the respective datasets. 
Hence, in this section, we test the \new{applicability} of the ELECTS-LSTM model with identical hyper-parameters to different datasets in Europe (France and Germany) in \cref{fig:datasets:europe} and Africa (Ghana and South Sudan) in \cref{fig:datasets:africa}.

We organize this section in two parts. First, we discuss the model performance on large-scale European datasets where several ten to hundred thousands of field annotations are available.
Then, we train and test the model on two African datasets where substantially less training data is available for end-to-end optimization of this deep learning model.

\subsubsection{Large-scale datasets in Europe}

\begin{figure*}
	\begin{subfigure}{.5\textwidth}
	    \centering\includegraphics[width=.75\textwidth]{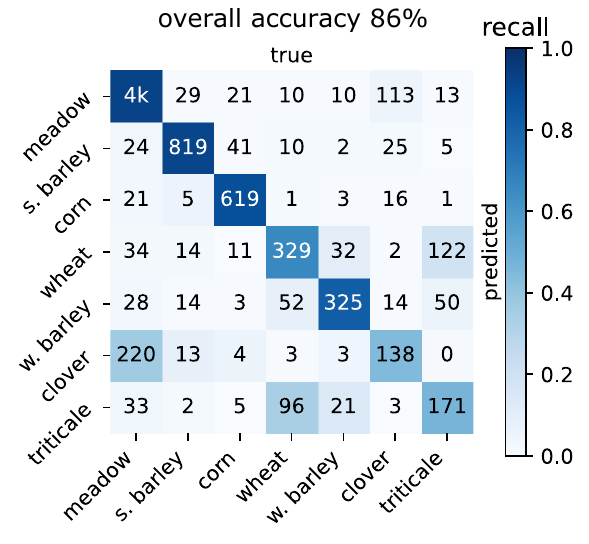}
	    \centering\includegraphics[width=.75\textwidth]{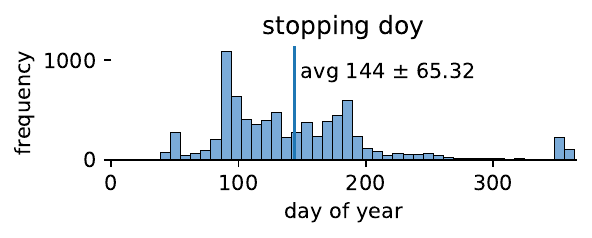}
	    \caption{BavarianCrops in Germany}
	    \label{fig:dataset:europe_bavariancrops}
	\end{subfigure}
	\begin{subfigure}{.5\textwidth}
	    \centering\includegraphics[width=.75\textwidth]{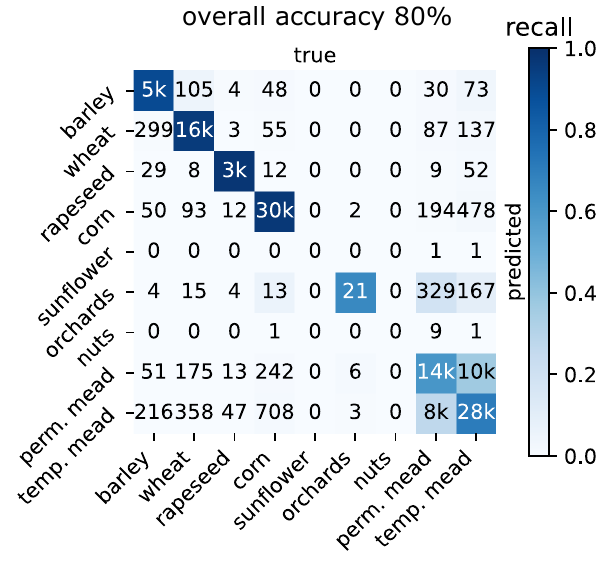}
	    \centering\includegraphics[width=.75\textwidth]{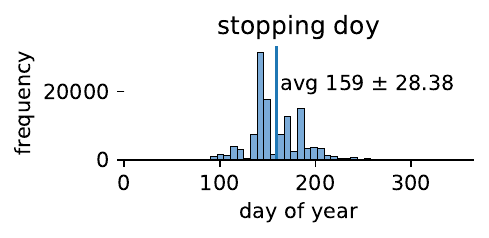}
	    \caption{BreizhCrops in Brittany, France}
	    \label{fig:dataset:europe_breizhcrops}
	\end{subfigure}
	
	\caption{Class-wise accuracy and earliness of the predictions on the two large-scale European datasets, BavarianCrops (a) and BreizhCrops (b). \new{The ELECTS model  predicts most classes accurately at a fraction of the required length of the time series in both datasets.}}

	\label{fig:datasets:europe}
\end{figure*}

 \cref{fig:datasets:europe} shows the accuracy as a confusion matrix and the earliness as a histogram of stopping times of field parcels in France (BreizhCrops) and Germany (BavarianCrops).
All crop classes within the BavarianCrops dataset (\cref{fig:dataset:europe_bavariancrops}) were classified accurately with an overall accuracy of 86\%. Systematic confusions were present between \class{wheat}, \class{winter barley}, and \class{triticale}, as these crops share biological ancestry and \class{clover} and \class{meadows} which are cultivated in a similar way and cut periodically throughout the year.
Most notably, the model achieves this accuracy with only 40\% of the entire sequence length. 
While the entire time series spans from January to December, most field parcels were classified within a two-month window (65 days) around May 24th. This highlights the potential of the ELECTS early classification approach to come to early and still accurate classification decisions within the year.

In \cref{fig:dataset:europe_breizhcrops}, we show the accuracy and earliness results on the BreizhCrops with fields of Brittany, France. 
Here, all crops are classified with an overall accuracy (OA) of 80\% with an average stopping period of one month around June 7th.
The ELECTS-LSTM model used only 32\% of the overall time series on average. Most notably, regular accuracy-only models, which make predictions at the end of the entire time series, achieve comparable accuracies of 80\% OA score, as shown in \cref{tab:breizhcrops_comparison} of \cref{appendix:comparison:breizhcrops}. In terms of classifications, systematic confusions are visible between \class{permanent meadows} and \class{temporary meadows}. Infrequent classes, such as \class{nuts} and \class{sunflowers} are not predicted correctly, as they have little effect on the overall loss objective. This classification of very imbalanced class distributions falls beyond the scope of this work.
Overall, these results show that the ELECTS-modified LSTM model matches the accuracy of regular non-early classification models while predicting substantially earlier within the season.

\subsubsection{Small-scale datasets in Africa}

\begin{figure*}
    \centering
    
	\begin{subfigure}{\textwidth}
  \hfill\includegraphics{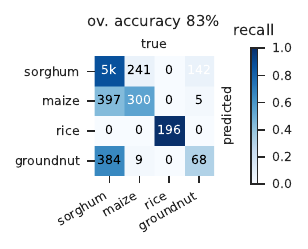}
 \hfill\includegraphics{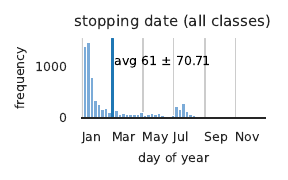}
\hfill\includegraphics{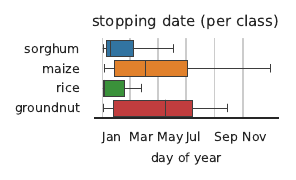}
\hfill
	    \caption{South Sudan}
	    \label{fig:datasets:africa:southsudan}
	\end{subfigure}
	
	\begin{subfigure}{\textwidth}
\hfill\includegraphics{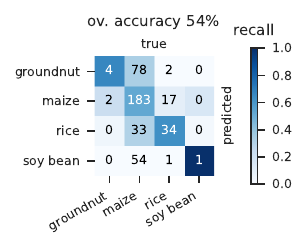}    
\hfill\includegraphics{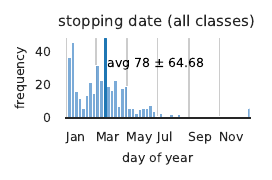}
\hfill\includegraphics{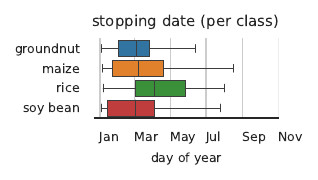}
\hfill
	    \caption{Ghana}
	    \label{fig:datasets:africa:ghana}
	\end{subfigure}
	
    \caption{Accuracy and Earliness on datasets in Africa from \cite{rustowicz2019semantic} that was recently integrated in the SustainBench dataset \cite{yeh2021sustainbench}. The trained ELECTS-LSTM finds a generally accurate and early solution for both datasets without changing the training configuration even though the dataset sizes are substantially smaller compared to the large-scale European datasets of France and Germany.}
    \label{fig:datasets:africa}
\end{figure*}

\Cref{fig:datasets:africa} shows confusion matrices and (class-wise) stopping times of the ELECTS-LSTM model in Ghana and South Sudan. 
The model finds accurate and early solutions on these datasets even though training a deep learning model on dataset sizes of \num{3837} and \num{737} individual field samples is inherently difficult. In South Sudan (\cref{fig:datasets:africa:southsudan}), an overall accuracy of 83\% is achieved with average predictions on the day of year 61 (March 2nd). These very early classifications are driven mainly by \class{rice} and \class{sorghum} fields that can be classified in January and February in this region.
This overall accuracy is on a similar level to a convolutional LSTM model with 82.6\% overall accuracy reported by Rustowicz \emph{et al.}, (2019) \cite{rustowicz2019semantic}. Note that their underlying classification model is advantaged: the kernels in their convolutional LSTM model can make use of the pixel-neighborhood. In comparison, the LSTM implementation, which we modified for ELECTS, can only classify individual pixels separately from each other. Note, however, that ELECTS can be modified to incorporate spatio-temporal data by changing the feature extractor to the same classifier as Rustowicz \emph{et al.}, (2019) \cite{rustowicz2019semantic}.
On a dataset of this comparatively small size, both deep learning models performed 5.7\% and 6.1\% worse compared to a regular random forest classifier (concatenates all time points to one large feature vector) with 88.7\% overall accuracy. 
Overall, the ELECTS-LSTM still compares well to the accuracy-only models from Rustowicz \emph{et al.}, (2019) \cite{rustowicz2019semantic} while only requiring a fraction of the time series to come to an accurate and early decision.
A similar trend is visible in the Ghana dataset shown in \cref{fig:datasets:africa:ghana} where the ELECTS-LSTM model achieves an overall accuracy of 54\% while classifying the fields on average on day of year 78 (March 19th). This accuracy is 7.1\% and 5.9\% worse compared to the random forest, and convolutional LSTM model from Rustowicz \emph{et al.}, (2019) \cite{rustowicz2019semantic}  that achieve 61.1\%, and 59.9\% accuracy, respectively. These accuracy-only models, however, can only predict after observing the entire time series, while the ELECTS-LSTM model used only 20\% of the overall time series with an average stopping date of the 78th day of the year. 

Overall, these results demonstrate that the ELECTS-LSTM model converges to a meaningful solution without any region-specific tuning. 
It produces early and still accurate predictions at a fraction of the entire time series. While the early classification model matched the accuracy of accuracy-only models on the large BreizhCrops dataset, the model achieved a marginally lower accuracy on the substantially smaller datasets in Africa. 
Still, early and accurate predictions have been achieved without any region-specific parameter tuning with the ELECTS-LSTM model, which demonstrates the applicability in unexplored areas where sufficient training data is available.

\section{Discussion}
\label{sec:discussion}

% following structure of https://www.sjsu.edu/writingcenter/docs/handouts/Discussion%20Section%20for%20Research%20Papers.pdf

%First, summarize the key findings from the research and link them to the initial research question. Seek to answer this question: What should readers take away from this paper?
%
In this work, we demonstrated that a regular long short-term memory (LSTM) recurrent neural network can be effectively modified with ELECTS for early and accurate classifications on various crop-type datasets. On sufficiently large datasets, the performance is on par with that of accuracy-only models with only a fraction of the sequence length.
%Furthermore, we associated the estimated stopping dates with local crop calendars. We showed that the ELECTS-LSTM model learned to utilize connections to physical events, such as the blossoming of the harvest of specific crops.
%
%Second, place the findings in context. This step will involve going back to the literature review section and analyzing how the results fit in with previous research. 
The ELECTS-LSTM does not require refitting or predicting with different sub-sequences, conversely to related work \cite{kondmann2022early,marszalek2020early,maponya2020pre} based on incremental classification. %It also provides  can only provide a class-wise estimate of achievable accuracy for a specific crop type. 
Other approaches \cite{lin2021early,konduri2020mapping} are often developed and targeted towards one specific deployment area, often focused on the continental US, while we demonstrated the applicability of ELECTS on different continents.
ELECTS inherits the limitations of deep learning: predominantly, the requirements of large annotated datasets for end-to-end optimization. %On the large-scale European datasets (BreizhCrops, and BavarianCrops), the ELECTS-LSTM model matched the accuracy of regular accuracy-only time series models. 
This was evident in the predictions on small datasets (Ghana and South Sudan) where the accuracy of the ELECTS-LSTM was marginally worse than an accuracy-only model, while it matched the accuracy at the European large-scale datasets. % that was hard-coded to predict at the end of the year. 
Sensitivity to label imbalance is a further limitation where wrong classifications of infrequent classes are penalized less than frequent ones.

Deploying an ELECTS model on applications beyond crop type mapping would be a natural extension, as this model can be trained on any class-annotated time series dataset. Extending ELECTS for spatio-temporal data is feasible with little effort and can be done by modifying the feature extractor, for instance, by adding 2D convolutional layers.   

The implications of an automated end-to-end trainable model, such as ELECTS, are manifold: 
acquiring predicted and accurate class labels for a subset of stopped crop parcels has direct practical implications for the control of European agricultural subsidies.
In practice, sample on-site inspections often control the  European subsidy after a specific pre-determined date. Field-wise, in-season predictions, \new{as ELECTS provides}, allow the start of this process weeks and months in advance.
Further, the potential to save computational and storage resources is substantial: ELECTS provided accurate predictions using between 16\% (South Sudan) to 40\% (BavarianCrops) of the overall time series. For instance, when scaling the average earliness of predictions in BavarianCrops to the 43TB of Sentinel-2 imagery acquired in Europe each year, 26TB of downloading and processing satellite can be avoided.
While Sentinel-2 data is free of charge, an increasing amount of daily high-resolution imagery is available today \cite{kondmann2022early}. This data needs to be acquired at a substantial cost and motivates the need to make confident decisions with data-efficient algorithms.
Training and deploying an ELECTS-LSTM model is not expensive in terms of computational efforts.
Deep learning models for 1D time series are small compared to standard 2D convolutional models for images. We trained the ELECTS-LSTM models within one hour on a single GPU on BavarianCrops. A researcher can make predictions using the trained ELECTS-LSTM model on a CPU with a regular notebook. 

\section{Conclusion}

We presented a training framework for End-to-end Learned Early Classification of Time Series (ELECTS) that augments a regular deep time series classification model by a second decision head informing about prediction uncertainty and leading to early stopping. The core contribution is a loss function that incorporates both model outputs such that the two objectives of earliness and accuracy are balanced.
Thanks to the earliness objective, ELECTS provides indirect insight into its decision process. We showed that the model linked the stopping decision to \new{the} phenological events of the plants for two crop types.
Stopped classifications early in the season were also particularly accurate, \new{highlighting} that the model connects the stopping decisions to predictive confidence.
ELECTS goes beyond crop types classification, as it can be applied to potentially any data where temporally coarse labels are available that are not aligned with the events, e.g., one label per year. 
In general, with satellites providing a constant stream of data that is necessary to monitor time-dependent processes at the surface~\cite{CampsValls21wiley}, a variety of deployments are feasible, from dynamically determining cloud categories~\cite{mateo2019convolutional} to the detection of deforested areas~\cite{reiche2021forest} in a time-sensitive manner. 
The source code to the models, the ELECTS loss function, and to reproduce the experiments are available at \texttt{https://github.com/marccoru/elects}.

\section{Acknowledgements}

The work of Marc Rußwurm was partially funded by the German Federal Ministry for Economic Affairs and Energy (BMWi) under reference 50EE1908. Romain Tavenard was partially funded through project MATS ANR-18-CE23-0006. Nicolas Courty is partially funded through the ANR project OTTOPIA ANR-20-CHIA-0030.

\bibliographystyle{unsrt}
\bibliography{bib/references.bib}

\appendix

\section{Ablation Experiments}
\label{appendix:ablations}

\subsection{Ablations on Hyperparameters and Loss Design}

In this group of experiments, we test the individual model components.
In \cref{sec:exp:alpha}, we vary the trade-off between accuracy and earliness while we focus in \cref{sec:ptsoffset} on the offset parameter $\varepsilon$.

\subsubsection{Controlling Earliness versus Accuracy}
\label{sec:exp:alpha}

\begin{table}

    \begin{subtable}{\linewidth}

        \centering
        \newcommand{\f}{\color{black}}

    	\centering\hspace{0em}\begin{tabular}{lcc|c}
    		\toprule\small
    		\textbf{$\alpha$} & accuracy& $\kappa$ & earliness \\
    		\cmidrule(lr){0-0}\cmidrule(lr){1-1}\cmidrule(lr){2-2}\cmidrule(lr){3-3}\cmidrule(lr){4-4}
    		0.0  &\f \f 0.25${\pm 0.22}$     &\f 0.12${\pm 0.19}$ &\f {0.90}${\pm 0.17}$      \\
    		0.2 & 0.81${\pm 0.03}$           & 0.71${\pm 0.04}$  & {0.60}${\pm 0.02}$  \\
    		0.4 & 0.80${\pm 0.09}$           & 0.71${\pm 0.10}$ & 0.53${\pm 0.03}$           \\
    		0.6 & {0.85}${\pm 0.02}$  & 0.77${\pm 0.03}$ & 0.12${\pm 0.07}$           \\
    		0.8 & 0.84${\pm 0.01}$           & 0.76${\pm 0.02}$ & 0.07${\pm 0.05}$           \\
    		1.0  &\f 0.83${\pm 0.03}$      &\f 0.75${\pm 0.04}$ &\f 0.00${\pm 0.00}$         \\
    		\bottomrule
    	\end{tabular}
    	\caption{BavarianCrops.}
    	\label{tab:alpha:bavaria}

    	\end{subtable}
     
    	\begin{subtable}{\linewidth}
    	    \centering\begin{tabular}{lcc|c}
    		\toprule\small
    		\textbf{$\alpha$} & accuracy& $\kappa$ & earliness \\
    		\cmidrule(lr){0-0}\cmidrule(lr){1-1}\cmidrule(lr){2-2}\cmidrule(lr){3-3}\cmidrule(lr){4-4}
    		$0.0$ & 0.31 & 0.00 & 1.00 ${\pm 0.00}$ \\
    		$0.2$ & 0.80 & 0.74 & 0.73 ${\pm 0.07}$ \\
    		$0.4$ & 0.80 & 0.74 & 0.69 ${\pm 0.07}$ \\
    		$0.6$ & 0.81 & 0.75 & 0.66 ${\pm 0.09}$ \\
    		$0.8$ & 0.80 & 0.74 & 0.60 ${\pm 0.12}$ \\
    		$1.0$ & 0.81 & 0.75 & 0.00 ${\pm 0.00}$ \\
    		\bottomrule
    	\end{tabular}
    	\caption{BreizhCrops.}
    	\label{tab:alpha:brittany}
    	\end{subtable}
    		\caption{Varying the weighting factor $\alpha$ that trades-off classification loss and earliness reward. An $\alpha=1$ corresponds to a high weight on earliness, while $\alpha=0$ switches off the earliness reward. Results for BavarianCrops are averaged over three runs. Standard deviations of earliness refer to stopping times of single fields.}
    	\label{tab:alpha}

\end{table}

In this experiment, we study the effect of $\alpha$ of \cref{eq:lcer} on both the BavarianCrops and BreizhCrops datasets.
For the BavarianCrops dataset, we observed some variance when training models from different weight initializations. Hence, we report the mean and standard deviation of $3$ model runs in \cref{tab:alpha:bavaria}.
For BreizhCrops, a similar accuracy level between $80$\% and $85$\% was achieved for a wide range of $\alpha$ values while the earliness decreased from 0.6 to 0.07. 
The accuracy-only ($\alpha=1$) runs did not achieve the best accuracy ($83$\%) compared to $85$\% with $\alpha = 0.6$. This result indicates that an earlier classification, temporally closer to the classification-relevant features, may have also been beneficial for the achieved accuracy.

These observations are mirrored in the BreizhCrops dataset in \cref{tab:alpha:brittany}. Consistent accuracy of $80$-$81$\%
is achieved for all $\alpha>0.2$. Similarly, larger weights on earliness reward with larger $\alpha$ values lead to slightly earlier classifications of $13$\% of the overall sequence length.
Comparing BreizhCrops and BavarianCrops, we observe that the accuracies in BreizhCrops were more consistent throughout the entire $\alpha$-range, which we associate with the $20$-times larger training set size. This larger quantity in labeled samples helps the model to find the optimum classification-relevant features in a certain time regardless of the model initialization and $\alpha$-weights.

\subsubsection{Effect of the Offset Parameter $\varepsilon$}
\label{sec:ptsoffset}

\begin{figure}
	\begin{subfigure}{.5\textwidth}
		\includegraphics[width=\textwidth]{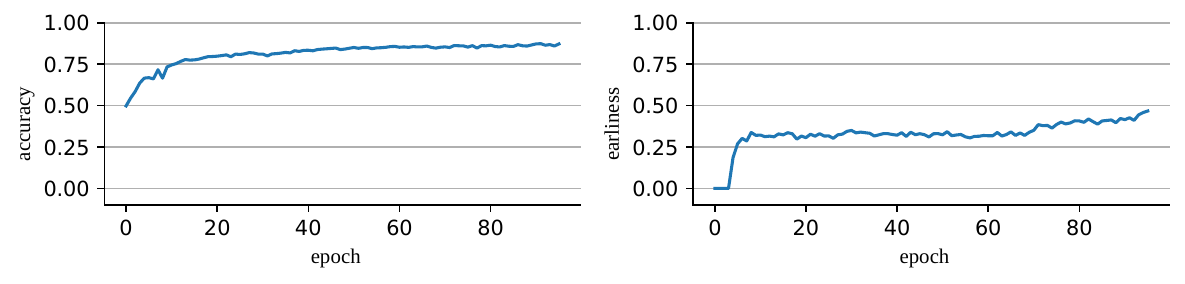}
		\caption{$\varepsilon = 10$. Normal training with early and accurate solutions. Observed in all $20$ runs with $\varepsilon=10$.}
		\label{fig:e10}
	\end{subfigure}

	\begin{subfigure}{.5\textwidth}
		\includegraphics[width=\textwidth]{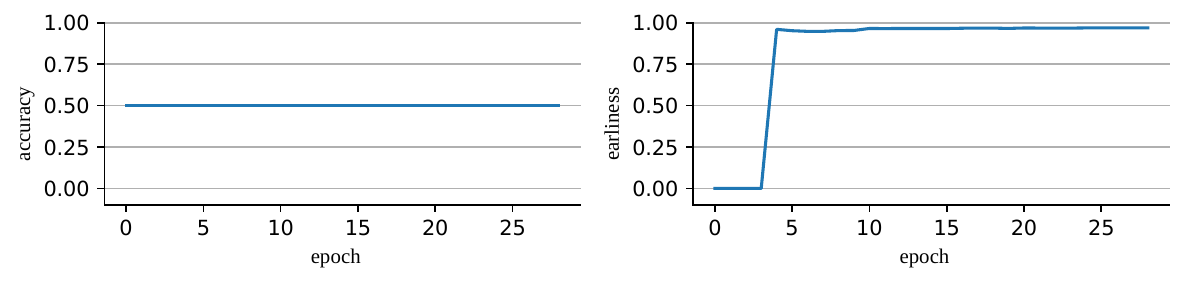}
		\caption{$\varepsilon = 0$. Failure case $1$.  Model only optimized earliness. Observed in $2$/$20$ runs.}
		\label{fig:e0f1}
	\end{subfigure}

	\begin{subfigure}{.5\textwidth}
		\includegraphics[width=\textwidth]{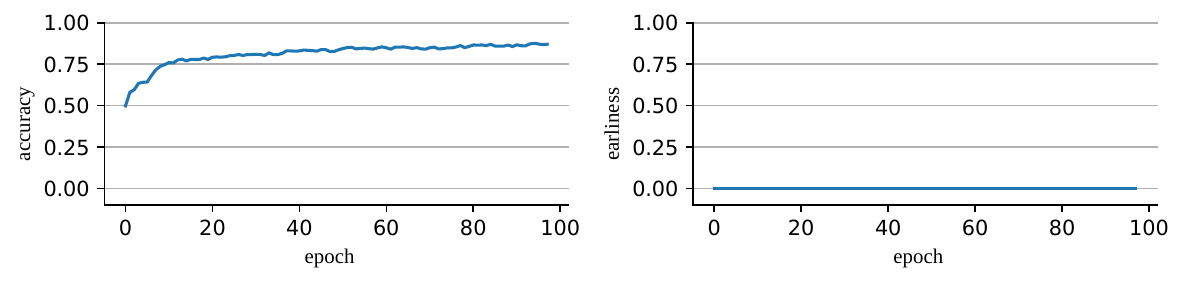}
		\caption{$\varepsilon = 0$. Failure case $2$. The  model only optimized for accuracy, which was observed in $1$/$20$ runs.}
		\label{fig:e0f2}
	\end{subfigure}

	\caption{We show normal training behavior (a) compared to two failure cases (b,c) we observed in some training runs with $\varepsilon=0$. In failure case $1$ (b), the model only optimized the earliness, while in failure case $2$ (c) only accuracy was optimized. None of the runs with $\varepsilon=10$ experienced these failure cases.}
	\label{fig:epsilon:qualitative}
\end{figure}

\begin{figure*}[t]
	\begin{subfigure}{\textwidth}
		\begin{tikzpicture}
\pgfplotstableread[col sep = comma]{images/satellite_comparison/ELECTS_unbalanced.csv}\elects
\pgfplotstableread[col sep = comma]{images/satellite_comparison/SR2CF2_unbalanced.csv}\mori

\begin{groupplot}[
group style={
	group size=3 by 1,
},
ymajorgrids,
legend columns=2,
width=.36\textwidth,
height=3cm,
xmode=log,
legend style={
    font=\tiny,
    at={(1,1.3)},
    },
ylabel style={yshift=-1.2em},
]

\nextgroupplot[
ylabel={accuracy},
xlabel={dataset size}]

\addplot[tumblue,mark=*] table[x=nsamples, y=accuracy]{\elects};
\addplot[tumorange,mark=*] table[x=nsamples, y=accuracy]{\mori};

\legend{ELECTS,SR2-CF2}

%\draw[tumgray] (axis cs:0,0.5566) -- (axis cs:10000,0.5566);
%\node[anchor=south, font=\tiny, text=tumgraydark, inner sep=0] at (axis cs:2100,0.5200){reference: predict most common class};

\nextgroupplot[
ylabel={earliness},
xlabel={dataset size},
xmode=log,
legend pos=south east,
ytick={0,0.25, 0.5,0.75, 1},
yticklabels={1,.75,0.5,.25,0}]

\addplot[tumblue,mark=*] table[x=nsamples, y=earliness]{\elects};
\addplot[tumorange,mark=*] table[x=nsamples, y=earliness]{\mori};

%\legend{ELECTS,SR2-CF2}

%\nextgroupplot[
%ylabel={earliness},
%xlabel={dataset size},
%xmode=log,
%legend pos=south east]
%
%\addplot[tumblue,mark=*] table[x=nsamples, y=score]{\elects};
%\addplot[tumorange,mark=*] table[x=nsamples, y=score]{\mori};
%
%\legend{ELECTS,SR2-CF2}

\nextgroupplot[
ylabel={runtime [h]},
xlabel={dataset size},
legend pos=north east,
ymode=log,
xmode=log,
ytick={1,10,100}]

\addplot[tumblue,mark=*] table[x=nsamples, y=total_runtimes]{\elects};
\addplot[tumorange,mark=*] table[x=nsamples, y=total_runtimes]{\mori};

\node[font=\tiny, anchor=west] (two) at (axis cs:1000,7.77) {7.77 h};
\node[font=\tiny, anchor=south, xshift=-0.5em] (two) at (axis cs:10000,0.105278) {6.31 min};

%\node[font=\tiny, anchor=north] (two) at (axis cs:2500,30) {30.25 h};
%\node[font=\tiny, anchor=south] (two) at (axis cs:2500,0.04) {2.90 min};

%\node[font=\tiny] (two) at (axis cs:5000,30) {104 h};
%\node[font=\tiny] (two) at (axis cs:5000,0.2) {3.58 min};

%\legend{ELECTS,SR2-CF2}

\end{groupplot}

\end{tikzpicture}
		\caption{unbalanced dataset}
		\label{fig:comparesat:unbalanced}
	\end{subfigure}

	\begin{subfigure}{\textwidth}
		\begin{tikzpicture}
%\pgfplotstableread[col sep = comma]{images/satellite_comparison/ELECTS_uniform.csv}\elects
%\pgfplotstableread[col sep = comma]{images/satellite_comparison/SR2CF2_uniform.csv}\mori

\begin{groupplot}[
group style={
	group size=3 by 1,
},
ymajorgrids,
legend columns=2,
width=.36\textwidth,
height=3cm,
ylabel style={yshift=-1.2em},
legend style={
    font=\tiny,
    at={(1,1.3)},
    },
]

\nextgroupplot[
ylabel={accuracy},
xlabel={dataset size}]

\addplot[tumblue,mark=*] table[col sep = comma, x=nsamples, y=accuracy]{images/satellite_comparison/ELECTS_uniform.csv};
\addplot[tumorange,mark=*] table[col sep = comma, x=nsamples, y=accuracy]{images/satellite_comparison/SR2CF2_uniform.csv};

\legend{ELECTS,SR2-CF2}

\nextgroupplot[
ylabel={earliness},
xlabel={dataset size},
legend pos=south east,
ytick={0,0.25, 0.5,0.75, 1},
yticklabels={0,.25,0.5,.75,1},
y dir=reverse,
y axis line style = {stealth-}]

% \draw[tumgray] (axis cs: 635.5,0) -- (axis cs: 635.5,1);
% \node[font=\scriptsize, text=tumgraydark](a) at (axis cs: 3500,.2){median size of UCR datasets};
% \draw[-Stealth, tumgray] (a) -- (axis cs: 635.5,.6);

\addplot[tumblue,mark=*] table[col sep = comma, x=nsamples, y=earliness]{images/satellite_comparison/ELECTS_uniform.csv};
\addplot[tumorange,mark=*] table[col sep = comma, x=nsamples, y=earliness]{images/satellite_comparison/SR2CF2_uniform.csv};

% \legend{ELECTS,SR2-CF2}

%\nextgroupplot[
%ylabel={earliness},
%xlabel={dataset size},
%legend pos=south east,
%ytick={0,0.5,1},
%yticklabels={1,0.5,0}]
%
%
%\draw[tumgray] (axis cs: 635.5,0) -- (axis cs: 635.5,1);
%\node[font=\tiny, text=tumgraydark](a) at (axis cs: 3500,.8){median size of UCR datasets};
%\draw[-Stealth, tumgray] (a) -- (axis cs: 635.5,.6);
%
%\addplot[tumblue,mark=*] table[x=nsamples, y=score]{\elects};
%\addplot[tumorange,mark=*] table[x=nsamples, y=score]{\mori};
%
%\legend{ELECTS,SR2-CF2}
%

\nextgroupplot[
ylabel={runtime [h]},
xlabel={dataset size},
legend style={at={(0,0.9)},anchor=west},
ymode=log,
ytick={1,10,100}]

\addplot[tumblue,mark=*] table[col sep = comma, x=nsamples, y=total_runtimes]{images/satellite_comparison/ELECTS_uniform.csv};
\addplot[tumorange,mark=*] table[col sep = comma, x=nsamples, y=total_runtimes]{images/satellite_comparison/SR2CF2_uniform.csv};

\node[font=\tiny, anchor=north] (two) at (axis cs:2500,30) {30.25 h};
\node[font=\tiny, anchor=south] (two) at (axis cs:2500,0.04) {2.90 min};

\node[font=\tiny] (two) at (axis cs:5000,30) {104 h};
\node[font=\tiny, xshift=-0.5em] (two) at (axis cs:5000,0.2) {3.58 min};

% \legend{ELECTS,SR2-CF2}

\end{groupplot}

\end{tikzpicture}
		\caption{balanced dataset}
		\label{fig:comparesat:balanced}
	\end{subfigure}
	\caption{This figure shows the evaluation of ELECTS (ours) and SR2-CF2 performance on class-balanced subsets of the BavarianCrops dataset. The x-axis refers to the size of the training data to train both models. We can observe that SR2-CF2 does not converge to a meaningful solution on imbalanced data (a) where it predicts at the beginning of the sequence (earliness = 0) at a low accuracy (as it could not observe any classification-relevant features that early). We needed to artificially balance the dataset in (b) for SR2-CF2 to predict early and accurately. In comparison, the ELECTS-LSTM converges to a meaningful solution in both cases and is computationally more efficient. With 5000 training time series in the balanced case, it required four minutes to train, while SR2-CF2 required 104 hours.}
	\label{fig:comparesat}
\end{figure*}

We trained ELECTS-LSTM $40$ times for $100$ epochs on the BavarianCrops dataset in this experiment. In $20$ training runs, we set $\varepsilon=0$ leading to no offset in \cref{eq:d}. In the second set of $20$ runs, we set $\varepsilon=10$. 
In $37$ of $40$ runs, we observed a normal training behavior, as shown in \cref{fig:e10} where classification accuracy and earliness increased throughout the training. All $20$ training runs with $\varepsilon=10$ showed this normal training behavior, while in $20$ runs with $\varepsilon=0$, we experienced two rare failure cases.
Two of twenty runs experienced failure case $1$, as shown in \cref{fig:e0f1}. In these runs, the earliness increased to $1$ early in the training leading to classification at the beginning of the sequence. The accuracy did not improve upon the initial epoch at $50$\%, which lies between the accuracy of a random predictor of $16$\% and predicting only the most frequent class (\class{meadow}) at an accuracy of $57$\%.
Here, the model fell in a local optimum where it solely minimized the earliness reward.
In \cref{fig:e0f2}, we show a second failure case that appeared one of twenty times on the runs with $\varepsilon=0$.
The accuracy increased steadily, but the predictions remained at the end of the sequence with an earliness of $0$.
In this case, the model minimized the classification loss but did not improve upon the earliness objective.
None of these failure cases were observed in the runs of $\varepsilon=10$ leading to a more stable convergence to an early and accurate solution with this offset parameter.

\section{Model Comparison}
\label{appendix:comparison}

In parallel, early classification has been discussed in the time series classification community, as summarized in the review of Gupta \textit{et al.}~\cite{gupta2020approaches}.
Here, early time series classification approaches are tested on a set of benchmark datasets in the UCR Archive \cite{UCRArchive2018}. While these datasets cover a diverse range of applications, their small size of at most a few thousand examples favors shallow learning solutions.
In this application space, several approaches introduced the idea to explicitly model the maximization of earliness in the optimization objective function \cite{dachraoui2015early,tavenard2016cost}.
In particular, Mori \textit{et al.}~\cite{mori2017early} also consider explicitly optimizing the trade-off between earliness and accuracy.
Their SR2-CF2 model variant first independently trains a Gaussian Process Classifier for each sub-sequence length. It then uses a genetic algorithm to find the parameter for a stopping rule that takes prediction confidence for each class into account.

\subsection{Comparison to SR2-CF2}
\label{appendix:comparison:mori}

In this first comparison, we use the BavarianCrops dataset.
The source code of SR2-CF2 was explicitly designed for uni-variate and class-balanced data in the UCR Time Series archive \cite{UCRArchive2018}.
We extended it to multi-variate time series in the modified source code\footnote{the extended source code to multi-variate time series of Mori \textit{et al.}, (2017) \cite{mori2017early} is available as fork \url{https://github.com/marccoru/earlyclassification}}.
We could not successfully run SR2-CF2 on the complete BavarianCrops dataset and sampled sub-datasets of \num{50}, \num{100}, \num{250}, \num{500}, \num{750}, \num{1000}, \num{2500}, \num{5000}, \num{7500}, and \num{10000} samples where we successfully ran SR2-CF2 on subsets up to \num{1000} samples.
Results are presented in \cref{fig:comparesat:unbalanced} where the SR2-CF2 method struggled to converge to a good solution. It predicted the most common class very late (small earliness) throughout differently-sized subsets of the crop type mapping dataset.
We connect this to the class imbalance present in the dataset. To alleviate this class imbalance, we sampled a second class-balanced dataset by undersampling frequent classes (e.g., \class{meadow}) and oversampling rare ones (e.g., \class{triticale}).
We created differently sized subsets of this balanced variant with \num{50}, \num{75}, \num{100}, \num{250}, \num{500}, \num{750}, \num{1000}, \num{2500}, \num{5000} samples and show the results in \cref{fig:comparesat:balanced}.
Here, the SR2-CF2 model achieved accurate and early classifications consistent with results reported on the UCR archives \cite{UCRArchive2018}. The ELECTS-trained model provided accurate but late (small earliness) classifications for datasets smaller than \num{2500} samples.
For datasets with \num{2500} training series or more, ELECTS and SR2-CF2 achieve comparable earliness, whereas the ELECTS-trained LSTM model predicted the classes more accurately.
While accuracy and earliness were generally comparable for datasets with more than \num{2500} samples, the difference in runtime, as shown in the last column, became a substantial factor.
The computational complexity of SR2-CF2 is $\mathcal{O}(N^2T^2)$ where $N$ refers to the number of samples in the dataset and $T$ to the sequence length.
The ELECTS-trained LSTM model relies on vanilla gradient descent that can utilize modern automatic differentiation libraries with a complexity of $\mathcal{O}(n_\text{epochs}NT)$.
In total, with a \num{5000} sample-sized dataset, ELECTS required $4$ minutes while SR2-CF2 $104$ hours.

\subsection{Comparison to non-early classification models on BreizhCrops}
\label{appendix:comparison:breizhcrops}

\begin{figure}[]
    \begin{subtable}{\linewidth}
    \small
    \centering
    \resizebox{\linewidth}{!}{
    \begin{tabular}{lrrrr}
    \toprule
    model & accuracy & kappa & earliness & average date of classification\\
    \cmidrule(lr){1-1}\cmidrule(lr){2-2}\cmidrule(lr){3-3}\cmidrule(lr){4-4}\cmidrule(lr){5-5}
    Random Forest & 0.78 & 0.69                               &  0 {\scriptsize(fixed)} &  Dec 28th {\scriptsize(fixed)} \\
    TempCNN \cite{pelletier2019temporal} & 0.79 & 0.73        &  0 {\scriptsize(fixed)} &  Dec 28th {\scriptsize(fixed)} \\
    MS-ResNet \cite{wang2018csi}  & 0.77 & 0.70               &  0 {\scriptsize(fixed)} &  Dec 28th {\scriptsize(fixed)} \\
    Inceptiontime \cite{fawaz2020inceptiontime} & 0.77 & 0.73 &  0 {\scriptsize(fixed)} &  Dec 28th {\scriptsize(fixed)} \\
    LSTM \cite{hochreiter1997long} & 0.80 & 0.74              &  0 {\scriptsize(fixed)} &  Dec 28th {\scriptsize(fixed)} \\
    Transformer \cite{vaswani2017attention} & 0.80 & 0.74     &  0 {\scriptsize(fixed)} &  Dec 28th {\scriptsize(fixed)} \\

    \midrule
    ELECTS-LSTM  & 0.80 & 0.74 & 0.68 $\pm$ 0.07 & \textbf{June 7th $\pm$ 28 days} \\

    \bottomrule

    \end{tabular}}
    
    \caption{Comparison with regular classification models on the BreizhCrops dataset.}
    \label{tab:breizhcrops_comparison}
    \end{subtable}
    \begin{subfigure}{\linewidth}
        \centering\includegraphics[width=\linewidth]{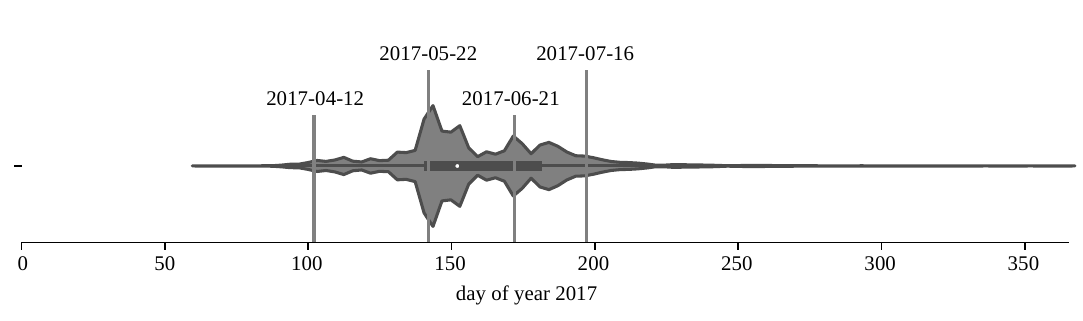}
        \caption{Frequency of dates where the classification has been stopped.}
        \label{fig:violin}
    \end{subfigure}
%    \begin{subfigure}{\textwidth}
%        \begin{tikzpicture}[node distance=0.2em]
%            \node[label=above:2017-04-12](a){\includegraphics[width=2.8cm]{images/frh04/20170412.png}};
%            \node[label=above:2017-05-22, right=of a](b){\includegraphics[width=2.8cm]{images/frh04/20170522.png}};
%            \node[label=above:2017-06-21, right=of b](c){\includegraphics[width=2.8cm]{images/frh04/20170621.png}};
%            \node[label=above:2017-07-16, right=of c](d){\includegraphics[width=2.8cm]{images/frh04/20170716.png}};
%        \end{tikzpicture}
%        \caption{Four Sentinel-$2$ scenes in a 2.5 km by 2.5 km site in the Brittany test region (2.4052° West, 47.5328° North).}
%        \label{fig:breizhcrops_comparison:qualitative}
%    \end{subfigure}
	\caption{Comparison of the ELECTS-LSTM model with other accuracy-only methods. ELECTS-LSTM matched the accuracy of the other models (a) while predicting at an earliness of $0.68 \pm 0.07$, meaning that only $32\% \pm 7\%$ of the time series was necessary. The average time of stopping is June $7$th $\pm~28$ days (earliness $0.68$ $\pm$ $0.07$) which lies in the greenup period in Brittany (b), as indicated by the four highlighted dates which correspond to the images (and analysis) in \cref{fig:maps}.}

\end{figure}

In this section, we compare the ELECTS-trained LSTM model with several models from the literature, optimizing for accuracy (Random Forest,
TempCNN \cite{pelletier2019temporal}, MS-ResNet \cite{wang2018csi}, Inceptiontime \cite{fawaz2020inceptiontime}, accuracy-only LSTM \cite{hochreiter1997long}, Transformer \cite{vaswani2017attention}). For such a comparison, we focus on the BreizhCrops benchmark.
The results are shown in \cref{tab:breizhcrops_comparison}.
The accuracy and kappa score measure the classification performance. In contrast, earliness $1-\frac{t}{T}$ measures how much data from the original T-length sequence was not needed to come to a prediction. The accuracy-only comparison models are always classified at the end of the sequence ($t=T$), which corresponds to a hard-coded earliness of $0$.
From \cref{tab:breizhcrops_comparison}, we see that the ELECTS-trained LSTM model matches the accuracies of the comparison models while predicting before the end of the sequence. But additionally to matching accuracy, ELECTS allows for earlier predictions (and the related savings in data download, storage, and processing time): in BreizhCrops, ELECTS achieves an earliness of $0.68$ $\pm$ $0.07$, meaning that only $32\% \pm 7$\% of the time series was necessary for the classification. This also means that the classification was stable (and stopped on June $7$th $\pm 28$ days rather than on December $28$ for the other methods, which need the entire time series.

Given that the evaluated earliness is early, we investigated the nature of the stopping period in greater detail in \cref{fig:violin}.
Here, we show the frequency of stopped dates of the ELECTS-LSTM model. We see that no classifications have been stopped before March (the 60th day of the year), which lies in the non-informative winter period. Most classifications have been made between the end of May and early June. %We investigate the stopping decisions in greater detail in the next section and ablation studies in the next section.

%
%\section{Algorithmic Representation}
%
%\begin{algorithm}
%	\caption{An algorithm with caption}\label{alg:cap}
%	\begin{algorithmic}
%		\Require $n \geq 0$
%		
%	\end{algorithmic}
%\end{algorithm}

\end{document}